\documentclass[10pt]{article}

\usepackage[T1]{fontenc}
\usepackage[utf8]{inputenc}
\usepackage{times}
\usepackage[margin=1in]{geometry}
\usepackage{amsmath,amssymb}
\usepackage{graphicx}
\usepackage{booktabs}
\usepackage{multirow}
\usepackage{makecell}
\usepackage{tabularx}
\usepackage{threeparttable}
\usepackage[table]{xcolor}
\usepackage{caption}
\usepackage{subcaption}
\usepackage{enumitem}
\usepackage{microtype}
\usepackage{tikz}
\usetikzlibrary{arrows.meta,positioning,fit,calc,shapes.geometric}
\usepackage[numbers,sort&compress]{natbib}
\usepackage[pdftitle={OceanTensorCube and ST-MMT: A Multimodal Data and Transformer System for Seven-Day Forecasting of Pyropia Bleaching},colorlinks=true,linkcolor=blue!50!black,citecolor=blue!50!black,urlcolor=blue!50!black]{hyperref}
\usepackage{authblk}
\usepackage{float}

\setlist{nosep,leftmargin=*}
\definecolor{cA}{HTML}{2a78d6}
\definecolor{cB}{HTML}{eb6834}
\definecolor{cC}{HTML}{1baf7a}
\definecolor{cD}{HTML}{eda100}
\definecolor{inkm}{HTML}{52514e}

\newcommand{\stmmt}{ST-MMT}
\newcommand{\otc}{OceanTensorCube}

\title{OceanTensorCube and ST-MMT: A Multimodal Data and Transformer System for Seven-Day Forecasting of \textit{Pyropia} Bleaching}

\author[1]{Gwangjin Kim\thanks{Corresponding author: \href{mailto:router128@hanmail.net}{router128@hanmail.net}}}
\author[2]{Cheolyoung Jang}
\author[3]{Yosep Jo}
\affil[1]{NCS Instructor, Smart Human Resources Development Institute, Republic of Korea}
\affil[2]{NH Networks LLC, Republic of Korea}
\affil[3]{DreamTech Co., Ltd., Republic of Korea}
\date{}

\begin{document}
\maketitle

\begin{abstract}
Bleaching (discoloration) of cultured laver (\textit{Pyropia} spp.) destroys the market value of a crop that is harvested from floating nets over a few months each winter, and the damage is usually noticed only after the blades have lost their pigment. We present \stmmt{}, a spatiotemporal multimodal transformer that forecasts bleaching seven days ahead for laver-farming areas on the south-western coast of Korea. The model reads a 14-day stack of 32 co-registered channels on a $64\times64$ grid at 30\,m resolution and combines two paths: a joint encoder over all channels, and three modality-specific encoders (in-situ ocean, atmosphere, satellite) that are fused by two rounds of multi-head cross-attention with the in-situ ocean stream as the query hub. The cross-attention output enters the joint representation through a learnable residual gate. Inputs are drawn from the \otc{}, a layered (Bronze/Silver/Gold) pipeline that aligns in-situ surveys, weather stations, ocean-model fields and satellite reflectance on a common grid. We also describe the Bleaching Label Calibrator (BLC), a procedure that converts point-in-time damage reports into daily soft labels and confidences; the model evaluated here is trained on binary labels. On a validation set of 2,792 cluster-day samples (176 positives, base rate 6.30\%), which was also used to select checkpoints and the alert threshold, a six-seed ensemble reaches ROC-AUC 0.9940 and PR-AUC 0.9097. At the operating threshold chosen to detect every reported positive (0.0940), precision is 65.7\% and the false-alarm rate is at most 3.5\% (92 of 2,616 negatives, all in 2024). Because positive labels occur only in October--November 2025, a calendar rule alone separates the labels. In a separate cross-validated training of the same architecture designed to remove this shortcut, the AUC among clusters on the same November dates was 0.976--0.986 (bootstrap lower bounds $\geq0.94$); the calendar independence of the deployed ensemble itself has not been measured. We also quantify the main operational obstacle: dissolved inorganic nitrogen (DIN) is surveyed only quarterly, and freezing the nutrient channels 30, 60 and 129 days before the input window lowers the AUC to 0.941, 0.833 and 0.527. We report the negative results that shaped the design, including a rejected proxy for nutrient drift, and list what remains unmeasured.
\end{abstract}

\section{Introduction}
\label{sec:intro}

Laver, the red alga known as \textit{gim} in Korea and \textit{nori} in Japan, is farmed on nets suspended in shallow coastal water and is one of the most valuable marine crops of East Asia~\citep{fao2024sofia,hwang2020seaweed,kim2017seaweed}. The cultivated species, formerly placed in \textit{Porphyra} and now largely assigned to \textit{Pyropia}~\citep{sutherland2011bangiales}, are tolerant of strong intertidal stress~\citep{blouin2011porphyra}, yet a recurring failure mode remains: the blades lose their dark pigment and turn yellow-brown or pale. This discoloration, which we call \emph{bleaching} following the Korean term \textit{hwangbaekhwa}, lowers product grade sharply. Long-term studies in the Seto Inland Sea and the Ariake Sea link it to the depletion of dissolved inorganic nitrogen (DIN), often after blooms of large diatoms such as \textit{Eucampia zodiacus} that compete with the farmed blades for nutrients~\citep{hori2008discoloration,nishikawa2011eucampia,ito2013eucampia,imai2006eutrophication}. Temperature also matters: low temperature has been shown to cause discoloration by repressing growth and nitrogen-transporter expression~\citep{takahashi2020low}, and nutrient-ratio limitation has been documented in Korean farming grounds~\citep{kim2019physiological}.

\begin{figure}[t]
\centering
\begin{subfigure}[b]{0.48\linewidth}
  \centering
  \includegraphics[width=\linewidth]{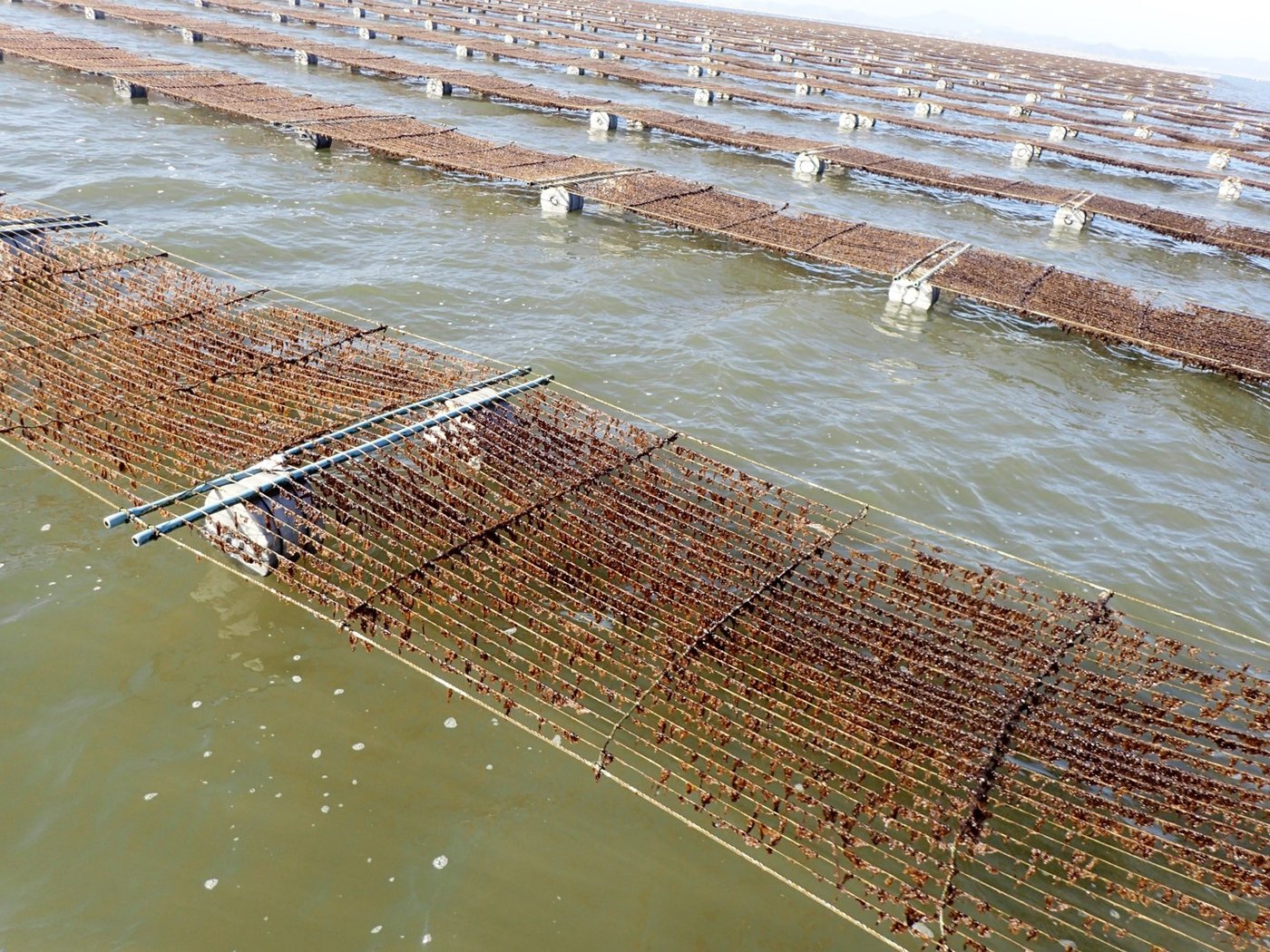}
  \caption{Laver on culture nets}
\end{subfigure}\hfill
\begin{subfigure}[b]{0.48\linewidth}
  \centering
  \includegraphics[width=\linewidth]{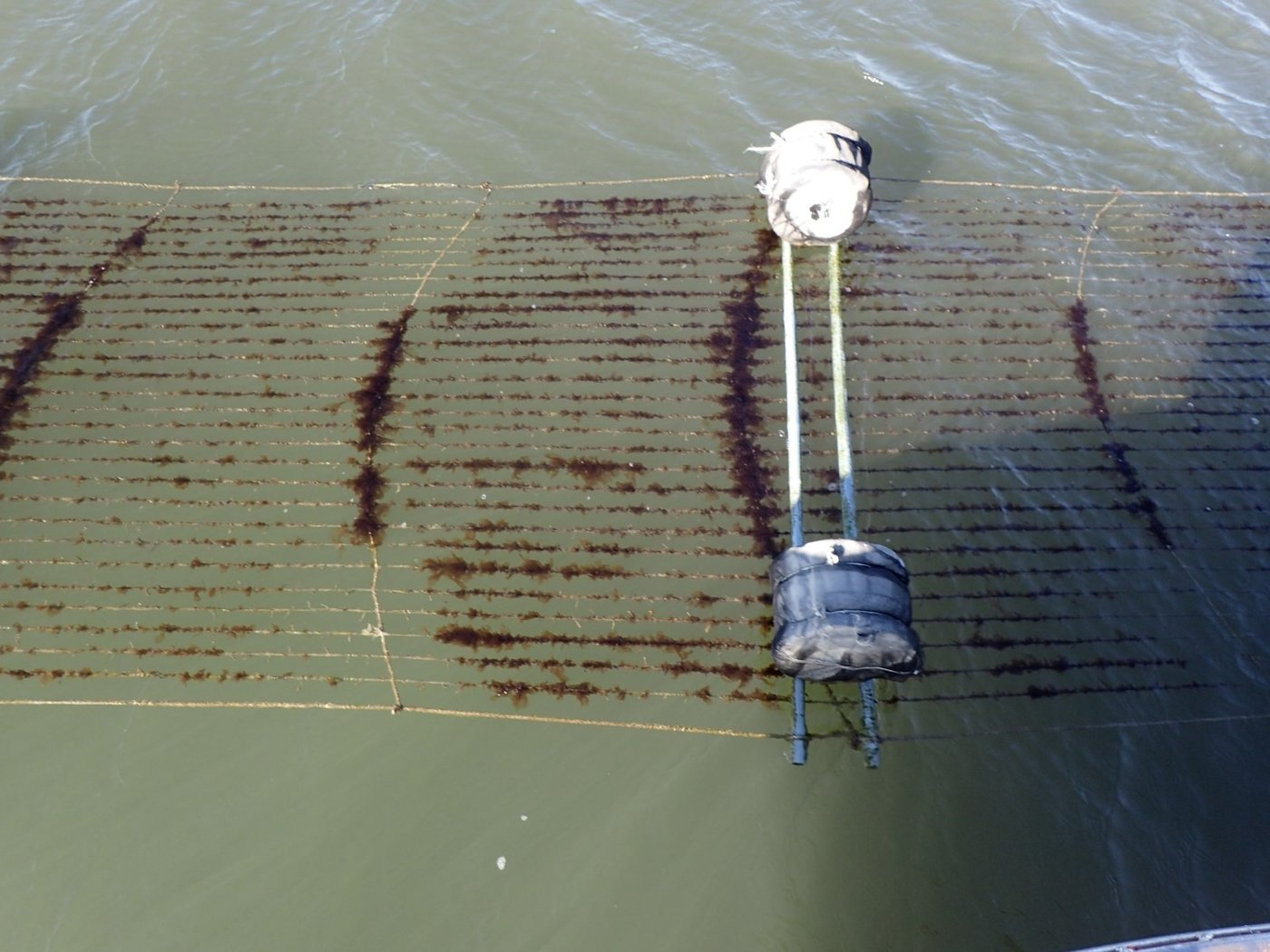}
  \caption{Net seen from above}
\end{subfigure}
\caption{Laver-farming grounds photographed in connection with the 2022 Haenam bleaching case. Photographs courtesy of the National Institute of Fisheries Science (NIFS).}
\label{fig:haenam}
\end{figure}

The 2022 event in Haenam, South Jeolla, shows the scale of a single episode (Figure~\ref{fig:haenam}). Bleaching began in early January 2022 at farms around Songji-myeon. By 4 February, a provisional count by the National Institute of Fisheries Science (NIFS) put 351 of the 418 laver-farming households in Haenam as affected, with 31,170 of the county's 70,208 net sets (one set measures 2.2\,m by 40\,m)~\citep{nifs2022haenamvisit}. On the same day a newspaper reported 2,236\,ha affected in the Songji, Hwasan and Hyeonsan townships, spreading toward Hwangsan, and losses of about 12.2 billion won~\citep{seoul2022haenam}. Samples from a joint survey on 18 January contained a bloom of the diatom \textit{Cylindrotheca closterium} at 1,653--9,958 cells/mL, which was presumed to have consumed nitrogen and other nutrients in the water~\citep{nifs2022haenamvisit}; NIFS later described the event as a case of nutrient depletion caused by mass growth of phytoplankton~\citep{nifs2022meeting}.

Farmers can respond to an expected nutrient shortage, for example by applying nutrients or by advancing the harvest, but only if they are warned before the pigment loss is visible. That makes bleaching a forecasting problem with three unusual properties.

\begin{enumerate}
  \item \textbf{Heterogeneous and asynchronous inputs.} The drivers are observed by different systems at very different cadences: hourly weather stations, daily ocean-model fields, weekly or 8-day satellite composites, and nutrient surveys that are repeated only a few times per year. A Korean national monitoring network measures dissolved inorganic nutrients four times a year~\citep{heo2011gunsan}.
  \item \textbf{Rare, report-based labels.} Bleaching is recorded when an agency or a farmer reports damage. The report date marks when the damage was noticed rather than when the environment turned unfavourable, and unreported cases (for example farms that recovered after timely fertilisation) are absent. The negatives are therefore \emph{unlabeled} rather than confirmed healthy, which is the positive--unlabeled setting~\citep{elkan2008learning,bekker2020learning}.
  \item \textbf{Asymmetric costs.} A missed event can mean a lost harvest, whereas a false alarm costs an unnecessary intervention. Operational evaluation should therefore fix the miss rate first and report the false alarms that remain.
\end{enumerate}

Machine learning has been applied to related coastal problems such as harmful algal bloom and red-tide prediction~\citep{cruz2021review,choi2022cnn,bak2019prediction,kim2019unet,hill2020habnet,lee2018improved} and sea-surface temperature forecasting on the same coast~\citep{kim2020sst}. Transformer architectures~\citep{vaswani2017attention} have become the standard tool for sequence and spatiotemporal data~\citep{dosovitskiy2021image,bertasius2021timesformer,arnab2021vivit,gao2022earthformer}, and cross-attention is the usual mechanism for fusing modalities~\citep{lu2019vilbert,tan2019lxmert,jaegle2021perceiver,chen2021crossvit}. We are not aware of a published deep-learning model that forecasts laver bleaching from multimodal ocean, weather and satellite data.

\paragraph{Contributions.} This paper describes a working system and is explicit about what has and has not been measured.
\begin{itemize}
  \item \textbf{\otc{}} (Section~\ref{sec:data}): a Bronze/Silver/Gold data pipeline that aligns in-situ nutrient surveys, weather stations, ocean-model currents and mixed-layer depth, and satellite near-infrared reflectance onto a common 30\,m grid, together with the rules we use to keep the inputs causal and symmetric between years.
  \item \textbf{BLC, the Bleaching Label Calibrator} (Section~\ref{sec:labels}): a procedure that turns point-in-time damage reports into daily soft labels, severity grades and confidences.
  \item \textbf{\stmmt{}} (Section~\ref{sec:model}): a residual architecture in which a joint spatiotemporal self-attention encoder is refined by hub-style multi-head cross-attention between modality groups.
  \item \textbf{An evaluation protocol for report-based labels} (Section~\ref{sec:protocol}): a recall-first operating threshold, sample-level counting, a same-date control against calendar shortcuts, and pre-registered acceptance criteria.
  \item \textbf{An operational analysis of the nutrient gap} (Section~\ref{sec:din}): how fast forecast skill decays when the nutrient survey ages, how persistence and climatology compare as stand-ins, and why a daily proxy for DIN drift was rejected.
\end{itemize}

\section{Related work}
\label{sec:related}

\paragraph{Laver discoloration and nutrients.}
Field studies in Japan established that nori discoloration follows declines in DIN and that winter diatom blooms are a frequent cause~\citep{hori2008discoloration,nishikawa2011eucampia,ito2013eucampia}. The broader context is a shift from eutrophic to oligotrophic conditions in enclosed seas~\citep{imai2006eutrophication}, which mechanistic ecosystem models try to reproduce~\citep{sohma2022modeling}. Physiological work identifies low temperature and nutrient-ratio limitation as further factors~\citep{takahashi2020low,kim2019physiological}, and disease must be distinguished from nutrient bleaching when yield falls~\citep{bae2025pyropia}. Our work is data-driven and complementary: it does not model the nitrogen budget but learns which combinations of observed conditions precede reported damage. The stoichiometric reasoning behind the nitrogen-to-phosphorus ratio channel follows \citet{redfield1958biological}.

\paragraph{Machine learning for coastal hazards.}
Reviews of harmful-algal-bloom forecasting describe a progression from regression and tree ensembles to deep networks that combine in-situ series and remote sensing~\citep{cruz2021review,hill2020habnet}. In Korea, convolutional networks driven by weather and ocean-model fields have predicted \textit{Cochlodinium} blooms in the South Sea~\citep{choi2022cnn}, ocean-colour imagery from the Geostationary Ocean Color Imager (GOCI)~\citep{ryu2012overview,choi2021goci2} has been used to detect red tides~\citep{kim2019unet}, imbalanced red-tide data have been addressed with oversampling~\citep{bak2019prediction}, and recurrent networks have forecast blooms in rivers~\citep{lee2018improved} and high water temperature near Goheung and Yeosu~\citep{kim2020sst}. Most of these studies treat events as reliably labeled and evaluate at a default threshold; our labels are sparser and noisier, which motivates the protocol in Section~\ref{sec:protocol}.

\paragraph{Attention for spatiotemporal and multimodal data.}
Multi-head attention~\citep{vaswani2017attention} applied to image patches~\citep{dosovitskiy2021image} and factorized over space and time~\citep{bertasius2021timesformer,arnab2021vivit} is the basis of our encoders. For weather and climate fields, Earthformer applies cuboid attention to Earth-system data~\citep{gao2022earthformer}. Time-series transformers such as Informer, the Temporal Fusion Transformer and PatchTST handle long horizons, mixed covariates and patching~\citep{zhou2021informer,lim2021temporal,nie2023patchtst}. Cross-attention fuses streams in vision--language models~\citep{lu2019vilbert,tan2019lxmert}, in Perceiver-style latent bottlenecks~\citep{jaegle2021perceiver,jaegle2022perceiverio} and between token streams of different scale~\citep{chen2021crossvit}. Convolutional recurrent models such as ConvLSTM~\citep{shi2015convolutional} and LSTMs~\citep{hochreiter1997long} remain strong baselines for gridded sequences.

\paragraph{Weak labels, imbalance and leakage.}
When negatives are only unlabeled, false-alarm rates measured against them are upper bounds and recall is measured only on reported events~\citep{elkan2008learning,bekker2020learning}. Under heavy class imbalance, precision--recall analysis is more informative than ROC analysis~\citep{davis2006relationship,saito2015precision}, and the Matthews correlation coefficient summarizes a confusion matrix more faithfully than accuracy or F1~\citep{chicco2020advantages}. Leakage, information in the inputs that would not be available at prediction time or that encodes the label indirectly, is a common cause of inflated results in applied machine learning~\citep{kaufman2012leakage,kapoor2023leakage}; Section~\ref{sec:protocol} describes the specific leakage paths we found and how we controlled them. We use focal loss~\citep{lin2017focal} for imbalance and seed ensembles~\citep{lakshminarayanan2017simple} for stability.

\section{Data: the \otc{}}
\label{sec:data}

\subsection{Study area and spatial units}
The unit of prediction is a \emph{tile cluster}. The national register of licensed laver-farming zones contains 1,194 polygons, concentrated on the coasts of South Jeolla (Jeonnam) and North Jeolla (Jeonbuk); we group them into 105 clusters. Each cluster is represented by a $512\times512$ tile at 30\,m resolution in the Korean projected coordinate system EPSG:5179, which is 15.36\,km on a side. The model sees the central $64\times64$ window (1.92\,km on a side, pixel indices 224--287 in both directions). Ten clusters without in-situ observations were excluded; patches exist for 81 clusters in 2024 and 95 in 2025. Forecasts are mapped back from clusters to farms; the current forecast pack covers 1,109 of the 1,194 farms, and the remaining 85 belong to clusters without 2024 or 2025 patches and are reported as ``no forecast'' rather than filled.

\subsection{Sources}
Table~\ref{tab:sources} lists the sources. They differ by more than two orders of magnitude in update frequency, which is the central difficulty of the problem. The marine environment monitoring network of the Korea Marine Environment Management Corporation (KOEM) provides nutrients and related water-quality variables at fixed stations; its nutrient survey campaigns are quarterly (February, May, August and November), consistent with the four-per-year schedule reported for the national network~\citep{heo2011gunsan}. After interpolating several stations onto a tile, the KOEM channels in our patches change value on 6--8 days per year, and one of our internal audits describes the cadence as six surveys per year; we use ``quarterly'' for the campaign schedule. Weather comes from the Korea Meteorological Administration (KMA) Automated Synoptic Observing System (ASOS)~\citep{kma_opendata}. Currents come from the HYCOM global ocean forecast system~\citep{chassignet2007hycom} for 2024 and from the Korea Hydrographic and Oceanographic Agency (KHOA) from 2025. Mixed-layer depth comes from an ocean-model product: our data audit traces it to the HYCOM mixed-layer thickness, while the latest build scripts read a daily CMEMS table~\citep{cmems2023globalphy}; we have not reconciled the two records. The near-infrared index used in the final patches comes from Sentinel-2 Level-2A 8-day composites extracted over farm polygons; earlier development versions used GOCI-II~\citep{ryu2012overview,choi2021goci2}. Daily biogeochemistry from the Copernicus Marine Service~\citep{cmems2023globalbgc} is ingested but is not used by the model reported here (Section~\ref{sec:history} describes a replacement experiment).

\begin{table}[t]
\centering
\caption{Data sources in the \otc{}. Cadence is the native update interval of the variables as used.}
\label{tab:sources}
\small
\begin{tabularx}{\linewidth}{@{}l X l l@{}}
\toprule
Source & Variables (channels) & Type & Cadence \\
\midrule
KOEM monitoring network & SST, DIN, DIP, SiO$_2$, salinity, chlorophyll, DO, pH, NO$_3$, NH$_4$, suspended solids (11) & point stations & quarterly surveys \\
KMA ASOS & precipitation, solar radiation, wind speed, wind direction (sin, cos), air temperature (6) & point stations & hourly $\rightarrow$ daily \\
HYCOM / KHOA; ocean model & surface current $u, v$; mixed-layer depth (3) & model grid / stations & sub-daily $\rightarrow$ daily \\
Sentinel-2 L2A & near-infrared index (1) & 30\,m imagery & 8-day composite \\
Derived & N:P ratio, SST anomaly, 7- and 30-day means, SST gradient, current speed, days since rain, NIR daily change (9) & computed & as parents \\
Calendar & month sine and cosine (2) & static & monthly \\
\midrule
NIFS reports & bleaching damage reports (labels) & events & irregular \\
\bottomrule
\end{tabularx}
\end{table}

\subsection{Layered pipeline}
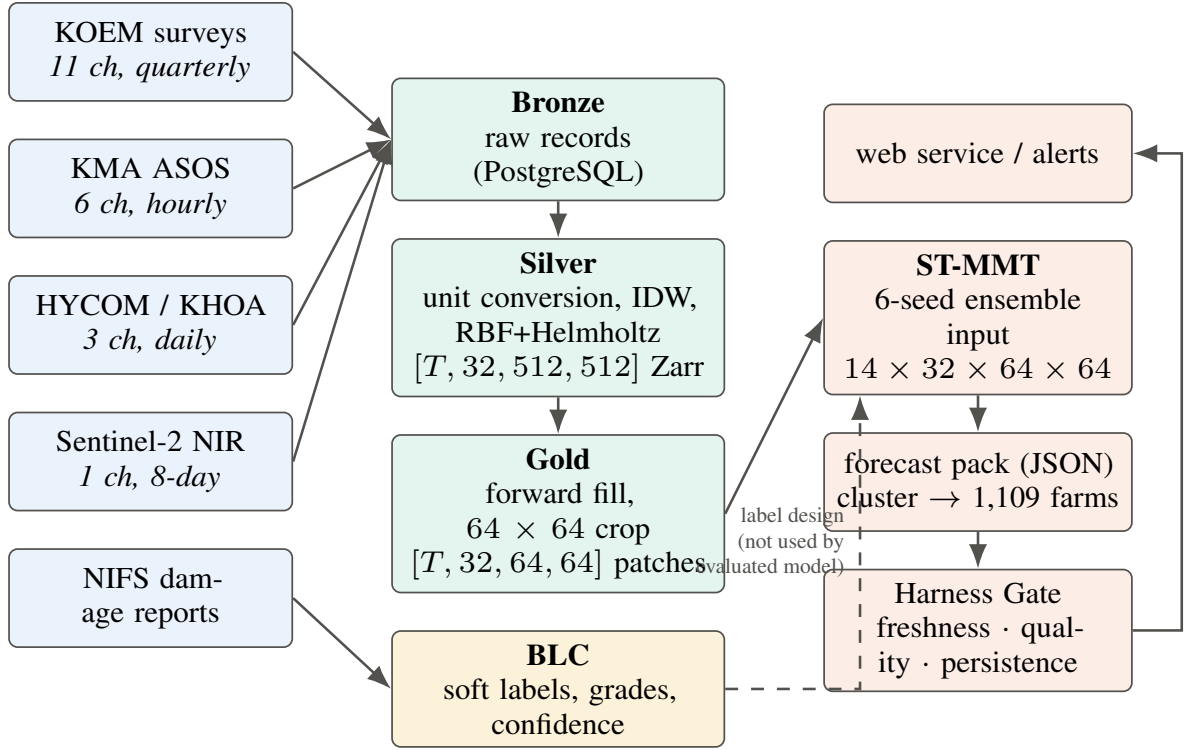
\begin{figure}[t]
\centering
\resizebox{0.95\linewidth}{!}{\begin{tikzpicture}[
  font=\scriptsize,
  box/.style={draw=inkm, rounded corners=2pt, align=center, minimum height=8mm, inner sep=3pt, fill=white},
  src/.style={box, fill=cA!10, text width=21mm},
  lay/.style={box, fill=cC!12, text width=25mm},
  srv/.style={box, fill=cB!12, text width=23mm},
  arr/.style={-{Latex[length=2mm]}, draw=inkm, line width=0.6pt},
  node distance=3mm and 6mm
]
\node[src] (koem) {KOEM surveys\\ \textit{11 ch, quarterly}};
\node[src, below=of koem] (kma) {KMA ASOS\\ \textit{6 ch, hourly}};
\node[src, below=of kma] (ocn) {HYCOM / KHOA\\ \textit{3 ch, daily}};
\node[src, below=of ocn] (sat) {Sentinel-2 NIR\\ \textit{1 ch, 8-day}};
\node[src, below=of sat] (rep) {NIFS damage reports};

\node[lay, right=8mm of kma, yshift=4mm] (bronze) {\textbf{Bronze}\\ raw records (PostgreSQL)};
\node[lay, below=of bronze] (silver) {\textbf{Silver}\\ unit conversion, IDW, RBF+Helmholtz\\ $[T,32,512,512]$ Zarr};
\node[lay, below=of silver] (gold) {\textbf{Gold}\\ forward fill, $64\times64$ crop\\ $[T,32,64,64]$ patches};
\node[lay, below=of gold, fill=cD!15] (blc) {\textbf{BLC}\\ soft labels, grades, confidence};

\node[srv, right=8mm of silver] (model) {\textbf{\stmmt{}}\\ 6-seed ensemble\\ input $14\times32\times64\times64$};
\node[srv, below=of model] (pack) {forecast pack (JSON)\\ cluster $\rightarrow$ 1,109 farms};
\node[srv, below=of pack] (gate) {Harness Gate\\ freshness $\cdot$ quality $\cdot$ persistence};
\node[srv, above=of model] (svc) {web service / alerts};

\foreach \s in {koem,kma,ocn,sat} \draw[arr] (\s.east) -- (bronze.west);
\draw[arr] (rep.east) -- (blc.west);
\draw[arr] (bronze) -- (silver);
\draw[arr] (silver) -- (gold);
\draw[arr] (gold.east) -- (model.west);
\draw[arr, dashed] (blc.east) -| node[pos=0.75, left, font=\tiny, inkm, align=right]{label design\\(not used by\\evaluated model)} ($(model.south west)+(3mm,0)$);
\draw[arr] (model) -- (pack);
\draw[arr] (pack) -- (gate);
\draw[arr] (gate.east) -- ++(4mm,0) |- (svc.east);
\end{tikzpicture}}
\caption{The \otc{} pipeline and serving path. Raw records (Bronze) are aligned to a 30\,m tile grid (Silver), forward-filled and cropped to $64\times64$ training patches (Gold). The six-seed ensemble writes a forecast pack that the web service reads through a three-stage output gate (Section~\ref{sec:gate}); the service never runs the network itself.}
\label{fig:pipeline}
\end{figure}

The pipeline follows a Bronze/Silver/Gold layering (Figure~\ref{fig:pipeline}).
\begin{itemize}
  \item \textbf{Bronze} stores raw records from each provider in a relational database, without transformation.
  \item \textbf{Silver} converts units, interpolates point data onto each tile grid and assembles a cube $\mathbf{V}\in\mathbb{R}^{T\times32\times512\times512}$ per tile and calendar year ($T=365$ or $366$). Nutrient concentrations reported in $\mu$g/L are converted to $\mu$mol/L by dividing by the atomic weight of the element (14.007 for nitrogen species, 30.974 for phosphate, 28.086 for silicate). Point observations are interpolated by inverse-distance weighting (power 2, nearest four stations for KOEM and nearest stations for KMA). Currents are interpolated with a thin-plate radial basis function and then projected onto their divergence-free part by a Helmholtz decomposition (1,500 Jacobi iterations), because a scalar interpolation of a vector field creates spurious convergence. Cubes are stored as Zarr arrays with one chunk per day $(1,32,512,512)$ and zstd compression; a raw tile-year of 12.25\,GB is stored in about 2.5\,GB.
  \item \textbf{Gold} forward-fills the sparse channels, extracts the central $64\times64$ patch, and writes one array of shape $(T,32,64,64)$ per cluster and year (tens of megabytes). Gold also stores the sample index and labels.
\end{itemize}
We use the name \otc{} for the resulting family of aligned tensors. The design target is a five-dimensional array indexed by region, month, channel and grid, of shape $[8, 72, 32, 512, 512]$ (eight regions, 72 months). What exists today are the per-tile, per-year four-dimensional slices described above.

\subsection{The 32 channels and their real information content}
Table~\ref{tab:channels} lists the channels. The count overstates the number of independent observations. The eleven KOEM channels come from a single water sample per survey, so they change on the same days: in the 2025 training patches the DIN channel has 7--9 distinct values per year and changes on only 6--8 days, and SST, salinity, chlorophyll, DO, pH, NO$_3$ and NH$_4$ change on exactly the same days. Counted by information source, the 32 channels comprise 6 daily weather variables, 3 daily ocean-model or current variables, 1 satellite variable, 11 low-frequency point variables, 9 derived variables and 2 calendar variables.

\begin{table}[t]
\centering
\caption{Channel definitions. Source is the source actually used by the extraction code; where it differs from the original channel master (turbidity, solar-radiation units, currents, mixed-layer depth), the code is authoritative.}
\label{tab:channels}
\scriptsize
\setlength{\tabcolsep}{3pt}
\begin{tabular}{@{}rlll@{\hspace{10pt}}rlll@{}}
\toprule
ch & Name & Source & Unit & ch & Name & Source & Unit \\
\midrule
0 & SST & KOEM & $^\circ$C & 16 & Wind speed & KMA & m/s \\
1 & DIN & KOEM & $\mu$mol/L & 17 & Wind direction (sin) & KMA & -- \\
2 & DIP & KOEM & $\mu$mol/L & 18 & Wind direction (cos) & KMA & -- \\
3 & SiO$_2$ & KOEM & $\mu$mol/L & 19 & Air temperature & KMA & $^\circ$C \\
4 & DIN/DIP ratio & derived & ratio & 20 & pH & KOEM & -- \\
5 & Salinity & KOEM & psu & 21 & NO$_3$ & KOEM & $\mu$mol/L \\
6 & Precipitation & KMA & mm/day & 22 & NH$_4$ & KOEM & $\mu$mol/L \\
7 & Solar radiation & KMA & MJ/m$^2$/day & 23 & SST gradient & derived & $^\circ$C/km \\
8 & Chlorophyll & KOEM & mg/m$^3$ & 24 & NIR index & Sentinel-2 & -- \\
9 & DO & KOEM & mg/L & 25 & Month (sin) & calendar & -- \\
10 & Current $u$ & HYCOM/KHOA & m/s & 26 & Month (cos) & calendar & -- \\
11 & Current $v$ & HYCOM/KHOA & m/s & 27 & Current speed & derived & m/s \\
12 & SST anomaly & derived & $^\circ$C & 28 & Chlorophyll 7-day mean & derived & mg/m$^3$ \\
13 & SST 7-day mean & derived & $^\circ$C & 29 & NIR daily change & derived & -- \\
14 & Days since rain & derived & days & 30 & SST 30-day mean & derived & $^\circ$C \\
15 & Suspended solids & KOEM & mg/L & 31 & Mixed-layer depth & HYCOM & m \\
\bottomrule
\end{tabular}
\end{table}

Two derived channels needed correction during development. The N:P ratio is computed as
\begin{equation*}
  \mathrm{N{:}P} = \mathrm{clip}\big(\mathrm{DIN}/\max(\mathrm{DIP},\,0.01),\;0,\;200\big);
\end{equation*}
an earlier form divided by $\mathrm{DIP}+10^{-6}$ and produced values up to $2.65\times10^{6}$ in one tile, which dominated the gradient saliency until it was fixed. The NIR channel 29 was first a 7-day moving average, which correlated with channel 24 at 0.78 and lowered AUC from 0.903 to 0.850 in a development run; the day-to-day change (correlation 0.26) restored it to 0.917, and the definition was changed accordingly.

\begin{figure}[t]
\centering
\includegraphics[width=0.66\linewidth]{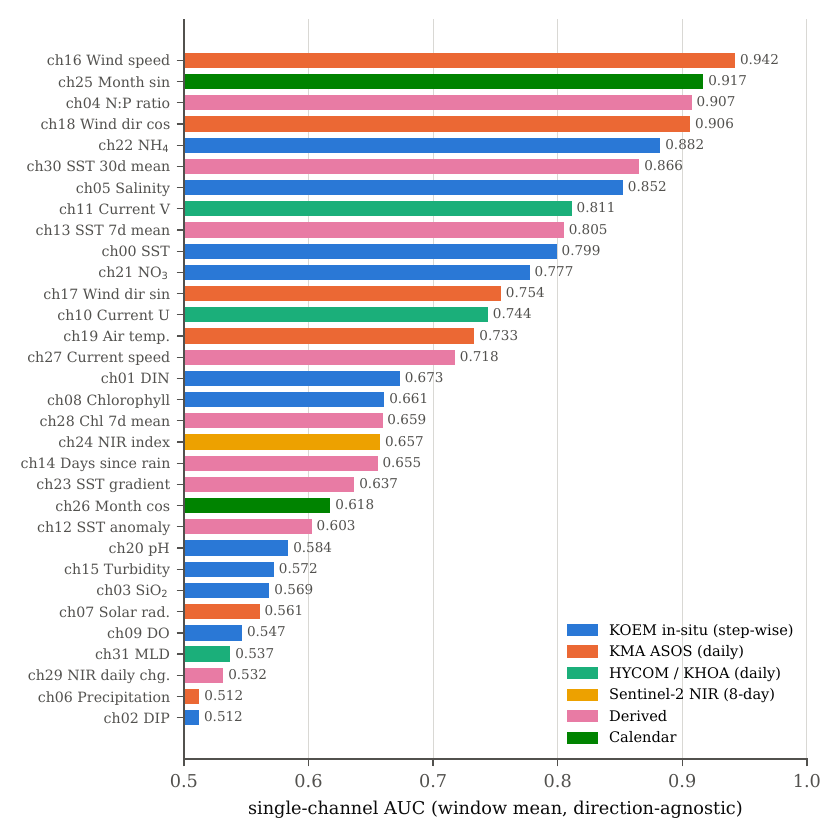}
\caption{Single-channel AUC of the 14-day window mean on the validation set (2,792 samples, 176 positives), coloured by information source. The calendar and season-correlated channels rank near the top, which motivates the same-date control in Section~\ref{sec:sdc}.}
\label{fig:chauc}
\end{figure}

Figure~\ref{fig:chauc} shows the discrimination of each channel on its own: the area under the ROC curve of the 14-day window mean, computed on the validation set and taken in whichever direction is larger. The strongest single channels are wind speed (0.942), month sine (0.917), N:P ratio (0.908) and the cosine of wind direction (0.906). For reference, the year alone scores 0.800 and the month alone 0.858. High single-channel scores for slowly varying, season-correlated variables are a warning sign rather than an achievement: they show how much of the label can be explained by when a sample was taken (Section~\ref{sec:labels-used}).

\subsection{Missing values, causality and year symmetry}
\label{sec:causality}
Three rules govern how gaps are filled. Each was introduced after a specific failure, and one known violation remains in the training data.

\paragraph{Forward fill.} Twelve sparse channels (the KOEM channels other than suspended solids, plus the N:P ratio and SST gradient) are forward-filled within each calendar year: an unobserved day carries the most recent observation. Forward filling is causal, but it encodes the assumption that the variable has not changed since the last survey (Section~\ref{sec:din}).

\paragraph{Missing as zero.} At patch extraction, remaining gaps are set to zero. Zero is a physically valid value for several channels (no rain, no sunlight), so the network cannot distinguish ``zero'' from ``unknown''. Section~\ref{sec:expA} measures what happens when channels that should carry values are zero at inference time.

\paragraph{Year symmetry.} Training contrasts a year without reported bleaching (2024) with a year with bleaching (2025), so any channel that exists in only one of the two years can identify the year and, through it, the label. The pipeline therefore writes values only for dates covered in both years and leaves the rest at zero in both. For example, the current record begins on 15 March, so the three current channels are zero from 1 January to 14 March in both years; in the training patches their zero fraction is exactly 0.200 (73 of 365 days), which confirms that the rule was applied. Daily biogeochemistry from CMEMS is available for both years but was not used.

\paragraph{A known violation: backward fill.} The Gold step also contains a second pass that fills the days before a year's first survey with that survey's value, which uses information from the future. This pass affects samples whose targets fall in January and February: 6,828 of the 30,576 samples (22.3\%). All positive labels fall in October and November, so the affected samples are negatives, but the inputs are not reproducible in real time. The operational path uses forward fill only; retraining without the backward pass has not yet been done.

\section{Labels}
\label{sec:labels}

\subsection{The Bleaching Label Calibrator}
Official bleaching records from the National Institute of Fisheries Science (NIFS) consist of a report date, a place name and a severity grade. We used 31 such reports from 2010 to 2025, obtained with the cooperation of the NIFS seaweed research division. Point-in-time reports are a poor training target for three reasons: the report date marks when a person noticed the damage, which can lag the onset by one to two weeks; a binary label cannot express gradual progression; and reports are too rare to train a model on their own. BLC converts each report into a spatiotemporal field of soft labels, severity grades and confidences. For a report $e$ at time $t_e$ and location $g_e$ it proceeds as follows.

\begin{enumerate}
  \item \textbf{Window.} Consider days $t\in[t_e-14,\,t_e+14]$ and pixels $p$ in a buffer around $g_e$. Place names are converted to coordinates, projected to EPSG:5179 and matched to the nearest grid cell; we recommend averaging over a $3\times3$ neighbourhood rather than using a single pixel.
  \item \textbf{Six-factor score.} Compute a weighted environmental anomaly score
  \begin{equation}
    s(t,p) = \sum_{k=1}^{6} w_k\, \mathbb{1}\big[\phi_k(t,p)\big],\qquad \sum_k w_k = 100,
  \end{equation}
  where the conditions $\phi_k$ and weights $w_k$ are listed in Table~\ref{tab:blc}.
  \item \textbf{Peak normalization and time decay.} Normalize by the maximum score in the window and multiply by a linear decay with a floor:
  \begin{equation}
    r(t,p) = \frac{s(t,p)}{\max_{t',p'} s(t',p')},\qquad
    \delta(t) = \max\!\Big(0.3,\; 1-\frac{|t-t_e|}{14}\Big),\qquad
    y(t,p) = r(t,p)\,\delta(t)\in[0,1].
  \end{equation}
  \item \textbf{Severity.} Grade 0 (normal) if $y<0.3$, 1 (caution) if $0.3\le y<0.5$, 2 (warning) if $0.5\le y<0.8$, and 3 (severe) if $y\ge0.8$.
  \item \textbf{Confidence.} $c(t)=0.95$ for $|t-t_e|\le3$, $0.85$ for $\le7$, $0.70$ for $\le14$, and $0.50$ outside the window.
  \item \textbf{Storage.} Write $(y, c, \text{grade})$ into a label cube aligned with the \otc{}. Outside all event windows, $y=0$ with low confidence.
\end{enumerate}

\begin{table}[t]
\centering
\caption{BLC six-factor weak-label score. Thresholds are relative to the regional normal; the weights are initial values intended to be tuned by ablation.}
\label{tab:blc}
\small
\begin{tabular}{@{}llp{7.3cm}r@{}}
\toprule
Factor & Channel & Condition $\phi_k$ & $w_k$ \\
\midrule
DIN deficit & ch1 & below the regional lower quartile, exponentially weighted over 10 days & 30 \\
DIP deficit & ch2 & below the regional lower quartile, exponentially weighted over 10 days & 15 \\
N:P imbalance & ch4 & $<10$ (nitrogen-limited) or $>30$ (phosphorus-limited) & 15 \\
SST anomaly & ch12 & outside $\pm2\,^\circ$C of normal during the growing season & 15 \\
Short-term chlorophyll drop & ch28 & 7-day mean at least 30\% below the 14-day mean & 15 \\
Rain deficit & ch14 & more than 14 days since rain & 10 \\
\bottomrule
\end{tabular}
\end{table}

\begin{figure}[t]
\centering
\includegraphics[width=0.78\linewidth]{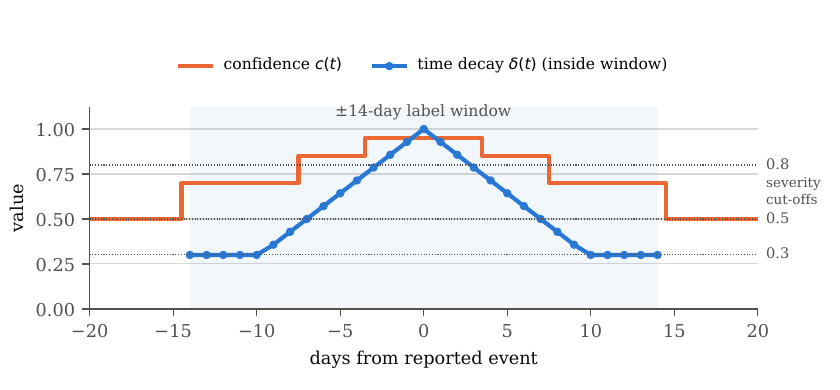}
\caption{BLC time rules around a reported event: the time-decay factor $\delta(t)$ that multiplies the normalized six-factor score inside the $\pm14$-day window, the confidence schedule $c(t)$, and the cut-offs that map the soft label $y$ to severity grades 0--3.}
\label{fig:blc}
\end{figure}

The BLC design motivates the $\pm14$-day window by three time scales: the one- to two-week lag between onset and report, an assumed one- to two-week physiological response of the blades, and the roughly two-week spring--neap tidal cycle. These are design assumptions rather than measured quantities. Figure~\ref{fig:blc} shows the decay and confidence schedules. BLC is designed to supervise a model through a confidence-weighted regression loss, $\mathcal{L}_{\text{main}} = \operatorname{mean}\big(c\cdot(\hat y-y)^2\big)$, with an optional masked-reconstruction auxiliary term weighted 0.3.

Applied to the 2024 Namhae coast as a first test (environmental scoring with event confidences), BLC produced 122 labels: 85 normal, 30 caution, 7 warning and none severe, with soft labels mostly between 0.1 and 0.4 and confidences of 0.70, 0.85 and 0.95. In development experiments a confidence-weighted loss was used with confidences assigned by evidence type (0.95 for official reports, 0.80 for field confirmation, 0.50 for environmental estimates) so that ``not reported'' is not learned as ``did not occur''.

\subsection{Labels used in this study}
\label{sec:labels-used}
The classifier evaluated in Section~\ref{sec:results} is trained and evaluated on binary cluster-day labels (version v7 of our label set), not on the BLC soft-label cube, and the training records do not state whether the final checkpoints used confidence weights. The positive days are recorded per cluster and date; our records do not trace each of them back to an individual report window. BLC is therefore presented as the labeling methodology of the platform, and its effect on forecast accuracy has not been isolated.

A sample is a pair (cluster, target date) with target dates from 21 January 2024 to 17 November 2025: 30,576 samples, of which 380 are positive (Table~\ref{tab:labelcomp}). Every positive lies in October or November 2025, and no negative does. The validation set consists of 8 clusters over both years, 2,792 samples with 176 positives (base rate 6.30\%); all 92 false alarms reported below come from 2024.

\begin{table}[t]
\centering
\caption{Composition of the labeled samples (left) and of the validation set (right).}
\label{tab:labelcomp}
\small
\begin{tabular}{@{}lrr@{\hspace{18pt}}lrrr@{}}
\toprule
\multicolumn{3}{c}{All samples} & \multicolumn{4}{c}{Validation (8 clusters $\times$ 2 years)} \\
\cmidrule(r{12pt}){1-3}\cmidrule(l){4-7}
Period & Positive & Negative & Year & Total & Positive & Negative \\
\midrule
2024, Jan--Dec & 0 & 18,144 & 2024 & 1,568 & 0 & 1,568 \\
2025, Jan--May & 0 & 12,052 & 2025 & 1,224 & 176 & 1,048 \\
2025, 1--20 Oct (2 clusters) & 40 & 0 & & & & \\
2025, 1--17 Nov (20 clusters) & 340 & 0 & & & & \\
\midrule
Total & 380 & 30,196 & Total & 2,792 & 176 & 2,616 \\
\bottomrule
\end{tabular}
\end{table}

This label structure has two consequences that shape the whole evaluation.
\begin{enumerate}
  \item \textbf{Calendar shortcut.} The single rule ``year is 2025 and month is October or November'' separates all 380 positives from all 30,196 negatives (AUC 1.000). A model can score highly by learning the calendar, either directly from channels 25--26 or indirectly from any seasonal variable. Section~\ref{sec:sdc} tests discrimination with the calendar held fixed.
  \item \textbf{Positive--unlabeled negatives.} NIFS seaweed researchers told us that farms showing early discoloration often recover after timely fertilisation, and such cases do not enter official statistics. Negatives are therefore ``not reported'', and the false-alarm rates we report are upper bounds; recall is measured only on reported events~\citep{elkan2008learning,bekker2020learning}.
\end{enumerate}

\section{Model: \stmmt{}}
\label{sec:model}

\subsection{Task and window arithmetic}
For a cluster $k$ and target date $\tau$, the model receives the 14 daily frames ending seven days before the target,
\begin{equation}
  \mathbf{X}_{k,\tau} = \mathbf{V}_k[\tau-20 : \tau-7] \in \mathbb{R}^{T\times C\times H\times W},\qquad T=14,\; C=32,\; H=W=64,
\end{equation}
where frame 0 is 1 January of the year, and outputs the probability $p_{k,\tau}$ that the cluster is in a bleaching state on day $\tau$. The forecast is issued on $\tau-7$, the last day of the input. We verified this arithmetic on every one of the 2,792 validation samples (input start $=\tau-20$, length 14, horizon 7, no exceptions). Getting it right was not trivial: in earlier development versions the target was set to the day after the input window, so those models were next-day predictors although they were described as seven-day predictors (Section~\ref{sec:history}).

\subsection{Architecture}
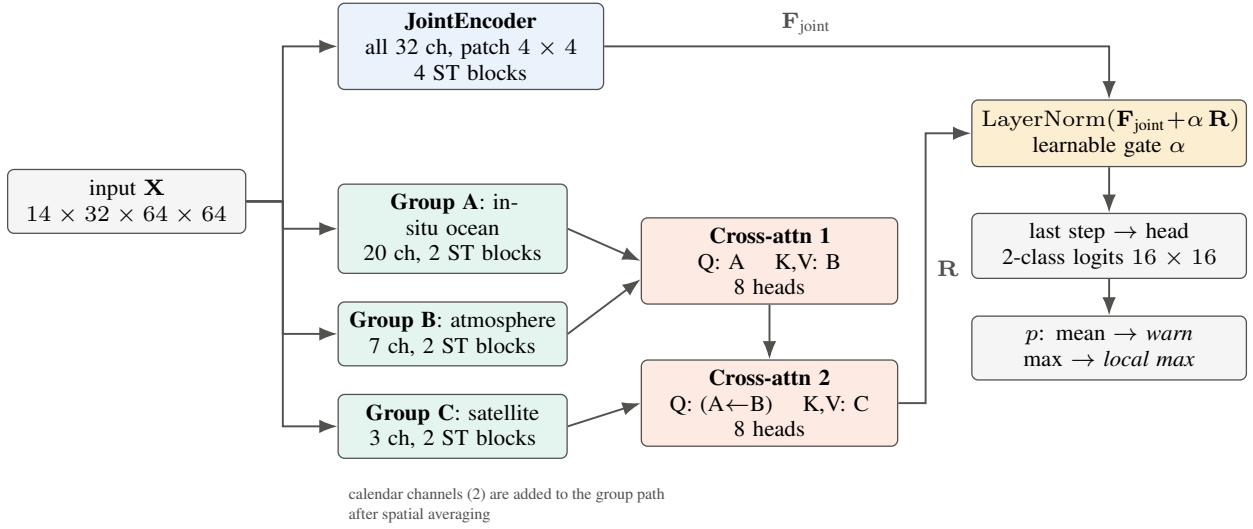
\begin{figure}[t]
\centering
\resizebox{\linewidth}{!}{\begin{tikzpicture}[
  font=\scriptsize,
  box/.style={draw=inkm, rounded corners=2pt, align=center, minimum height=7mm, inner sep=3pt},
  enc/.style={box, fill=cA!10, text width=27mm},
  grp/.style={box, fill=cC!12, text width=23mm},
  xa/.style={box, fill=cB!14, text width=26mm},
  arr/.style={-{Latex[length=2mm]}, draw=inkm, line width=0.6pt},
  node distance=4mm and 7mm
]
\node[box, fill=gray!8, text width=24mm] (in) {input $\mathbf{X}$\\ $14\times32\times64\times64$};

\node[enc, right=10mm of in, yshift=17mm] (joint) {\textbf{JointEncoder}\\ all 32 ch, patch $4\times4$\\ 4 ST blocks};

\node[grp, right=10mm of in, yshift=-3mm] (ga) {\textbf{Group A}: in-situ ocean\\ 20 ch, 2 ST blocks};
\node[grp, below=3mm of ga] (gb) {\textbf{Group B}: atmosphere\\ 7 ch, 2 ST blocks};
\node[grp, below=3mm of gb] (gc) {\textbf{Group C}: satellite\\ 3 ch, 2 ST blocks};

\node[xa, right=8mm of gb, yshift=8mm] (x1) {\textbf{Cross-attn 1}\\ Q: A \quad K,V: B\\ 8 heads};
\node[xa, below=6mm of x1] (x2) {\textbf{Cross-attn 2}\\ Q: (A$\leftarrow$B) \quad K,V: C\\ 8 heads};

\node[box, fill=cD!18, text width=28mm, right=8mm of x1, yshift=14mm] (fuse) {$\mathrm{LayerNorm}(\mathbf{F}_{\text{joint}} + \alpha\,\mathbf{R})$\\ learnable gate $\alpha$};
\node[box, fill=gray!8, text width=28mm, below=5mm of fuse] (head) {last step $\rightarrow$ head\\ 2-class logits $16\times16$};
\node[box, fill=gray!8, text width=28mm, below=4mm of head] (out) {$p$: mean $\rightarrow$ \textit{warn}\\ max $\rightarrow$ \textit{local max}};

\draw[arr] (in.east) -- ++(4mm,0) |- (joint.west);
\draw[arr] (in.east) -- ++(4mm,0) |- (ga.west);
\draw[arr] (in.east) -- ++(4mm,0) |- (gb.west);
\draw[arr] (in.east) -- ++(4mm,0) |- (gc.west);
\draw[arr] (ga.east) -- (x1.west);
\draw[arr] (gb.east) -- ($(x1.west)+(0,-2mm)$);
\draw[arr] (x1.south) -- (x2.north);
\draw[arr] (gc.east) -- (x2.west);
\draw[arr] (x2.east) -- ++(3mm,0) |- node[pos=0.25, right, inkm]{$\mathbf{R}$} (fuse.west);
\draw[arr] (joint.east) -| node[pos=0.2, above, inkm]{$\mathbf{F}_{\text{joint}}$} (fuse.north);
\draw[arr] (fuse) -- (head);
\draw[arr] (head) -- (out);
\node[inkm, align=left, anchor=north west, font=\tiny] at ($(gc.south west)+(0,-2mm)$) {calendar channels (2) are added to the group path\\ after spatial averaging};
\end{tikzpicture}}
\caption{\stmmt{} architecture (\texttt{STMMTCrossAttnV3}). A joint spatiotemporal encoder over all channels is refined by modality-specific encoders fused through two rounds of multi-head cross-attention with the in-situ ocean group as the query hub. The refinement enters through a learnable gate $\alpha$. The head outputs a $16\times16$ map of class probabilities; the forecast pack stores its spatial mean and maximum.}
\label{fig:arch}
\end{figure}

\stmmt{} has two paths that see the same input (Figure~\ref{fig:arch}). The joint path processes all 32 channels together and is the backbone. The group path splits the channels by physical origin, encodes each group separately, and fuses the groups with cross-attention. The group result refines the backbone through a gated residual connection. The design was chosen after a cross-attention-only variant underperformed a joint self-attention model at the recall-100\% operating point (Section~\ref{sec:history}): separating channels into groups before any joint processing prevents the encoder from capturing interactions across groups at the first layer, while a purely joint model has no explicit mechanism for one modality to query another.

\paragraph{Patch embedding.} Each frame is embedded by a strided convolution with kernel and stride 4,
\begin{equation}
  \mathbf{Z}_t = \mathrm{Conv2d}_{C\rightarrow d}^{\,4\times4,\;s=4}(\mathbf{X}_t) \in \mathbb{R}^{N\times d},\qquad N = (64/4)^2 = 256,\; d=256,
\end{equation}
so each token covers $4\times4$ pixels (120\,m $\times$ 120\,m). Group encoders use the same embedding on their channel subsets.

\paragraph{Spatiotemporal blocks.} An ST block applies spatial self-attention among the $N$ tokens of each frame, then temporal self-attention along the $T$ frames for each token position, then a position-wise feed-forward network, in the factorized manner of video transformers~\citep{bertasius2021timesformer,arnab2021vivit}. Each attention is the multi-head scaled dot-product attention of \citet{vaswani2017attention} with $h=8$ heads:
\begin{align}
  \mathrm{Attn}(\mathbf{Q},\mathbf{K},\mathbf{V}) &= \mathrm{softmax}\!\Big(\frac{\mathbf{Q}\mathbf{K}^{\top}}{\sqrt{d_h}}\Big)\mathbf{V},\qquad d_h = d/h = 32,\\
  \mathrm{MHA}(\mathbf{U},\mathbf{W}) &= \big[\mathrm{Attn}(\mathbf{U}\mathbf{W}^Q_i,\,\mathbf{W}\mathbf{W}^K_i,\,\mathbf{W}\mathbf{W}^V_i)\big]_{i=1}^{h}\,\mathbf{W}^O .
\end{align}
Self-attention is $\mathrm{MHA}(\mathbf{U},\mathbf{U})$. The joint encoder stacks four ST blocks; each group encoder stacks two.

\paragraph{Modality groups.} The 32 channels are divided as follows. Group A (in-situ ocean, 20 channels) contains the KOEM channels except suspended solids, the two current components and eight ocean-derived channels (N:P ratio, SST anomaly, SST 7-day and 30-day means, SST gradient, current speed, chlorophyll 7-day mean, mixed-layer depth). Group B (atmosphere, 7 channels) contains the six KMA channels and days since rain. Group C (satellite, 3 channels) contains suspended solids, the NIR index and its daily change; suspended solids sits here because the channel master originally assigned it a satellite source, although the code reads KOEM data (Table~\ref{tab:channels}). The two calendar channels are spatially averaged and added to the group representation, in the manner of a positional encoding.

\paragraph{Hub cross-attention.} Let $\mathbf{F}_A,\mathbf{F}_B,\mathbf{F}_C\in\mathbb{R}^{T\times N\times d}$ be the group encodings. The group path applies two cross-attention blocks with Group A as the query hub,
\begin{equation}
  \mathbf{H}_1 = \mathrm{CA}(\mathbf{F}_A;\,\mathbf{F}_B),\qquad
  \mathbf{R} = \mathrm{CA}(\mathbf{H}_1;\,\mathbf{F}_C),\qquad
  \mathrm{CA}(\mathbf{U};\mathbf{W}) = \mathrm{MHA}(\mathbf{U},\mathbf{W}),
\end{equation}
so the queries come from the in-situ ocean representation and the keys and values from the atmosphere and then from the satellite stream. In the implementation the attention weights have shape $(B\cdot T, N, N)$: at each time step, every ocean token attends over all spatial tokens of the other modality. Group A is the hub because in a gradient-saliency analysis of an earlier self-attention model most of the highest-ranked channels (precipitation, SST anomaly, N:P ratio, solar radiation, SST) belonged to Group A or to the closely related Group B. Fully pairwise bidirectional cross-attention among the three groups was tried and performed worse (Section~\ref{sec:history}).

\paragraph{Gated residual fusion and head.} The two paths are combined as
\begin{equation}
  \mathbf{F} = \mathrm{LayerNorm}\big(\mathbf{F}_{\text{joint}} + \alpha\,\mathbf{R}\big),
\end{equation}
with a learnable scalar $\alpha$ initialized at 0.1. If the refinement is unhelpful, training can drive $\alpha$ toward zero and the model reduces to the joint backbone. The representation at the last time step is passed to a classification head with two outputs per token, giving a $16\times16$ map of logits\footnote{The handover documentation of the deployed model describes a $16\times16$ probability map; an earlier report gives $64\times64$ output logits for the first cross-attention variant, whose decoder the residual model reuses. We have not reconciled the two descriptions.}; the positive-class probability is $p=\mathrm{softmax}(\cdot)_1$. The forecast pack stores the spatial mean of $p$ over the cluster window as the risk score and the spatial maximum as the local maximum risk.

\paragraph{Size.} With $d=256$, $h=8$, four joint and two group layers per group, and patch size 4, the model has 9,218,179 parameters (a 37\,MB checkpoint per seed). For comparison, the self-attention-only predecessor had 3,819,650 parameters and the first cross-attention-only variant 5,922,690.

\subsection{Training}
The loss is the focal loss~\citep{lin2017focal}
\begin{equation}
  \mathcal{L}_{\text{FL}} = -\alpha_{y}\,(1-p_{y})^{\gamma}\,\log p_{y},\qquad \gamma=2,\quad (\alpha_0,\alpha_1)=(0.5,\,1.5),
\end{equation}
where $p_y$ is the predicted probability of the true class, applied per output token, plus a spatial smoothness term on the output map; the class weights act as the focal $\alpha$ term. Gradients are clipped to norm 1.0. Inputs are standardized per channel with means and standard deviations computed on the training patches (standard deviations below $10^{-6}$ are set to 1), and the statistics are stored with each checkpoint. The checkpoint is selected by F1 at the recall-100\% threshold on the validation set, with early stopping (patience 10 for the adopted seven-day configuration). The optimizer, learning-rate schedule, batch size and the weight of the smoothness term are not recorded in the documents from which this paper was prepared, and we do not report them.

Six models are trained from different initializations (seeds 42, 1, 2, 3, 5 and 7), and the ensemble probability is the mean of their positive-class probabilities~\citep{lakshminarayanan2017simple}. The seeds differ substantially at the operating point: individual recall-100\% thresholds range from 0.0741 to 0.1247 and the false-alarm rate in 2024 from 5.5\% to 18.0\% among the seeds for which it was recorded, against 5.9\% for the ensemble. Averaging is therefore part of the method, not a refinement.

\begin{table}[t]
\centering
\caption{\stmmt{} configuration as deployed (\texttt{xattn\_v3\_focal\_7day}).}
\label{tab:hparams}
\small
\begin{tabular}{@{}ll@{\hspace{24pt}}ll@{}}
\toprule
Input & $14\times32\times64\times64$ & Hidden size $d$ & 256 \\
Horizon & 7 days & Attention heads & 8 \\
Patch size & $4\times4$ (256 tokens) & Joint / group ST blocks & 4 / 2 \\
Groups (A/B/C) & 20 / 7 / 3 channels (+2 calendar) & Cross-attention & A$\leftarrow$B, then $\cdot\leftarrow$C \\
Loss & focal, $\gamma=2$, class weights 0.5/1.5 & Fusion & $\mathrm{LN}(\mathbf{F}_{\text{joint}}+\alpha\mathbf{R})$ \\
Ensemble & 6 seeds, mean probability & Parameters & 9,218,179 \\
\bottomrule
\end{tabular}
\end{table}

\section{Evaluation protocol}
\label{sec:protocol}

\paragraph{Sample-level counting.} Every metric counts samples (cluster-days), not pixels. Several of our early development reports counted pixels in the output map, which multiplied each sample by up to 4,096 and made small validation sets look large; those numbers are not comparable with the ones reported here.

\paragraph{Recall-first operating threshold.} Because a missed event is the costlier error, the operating threshold is the highest threshold that still detects every reported positive in the validation set,
\begin{equation}
  \theta^{\star} = \max\{\theta : \mathrm{recall}(\theta)=1\} = \min_{i:\,y_i=1} \bar p_i - \epsilon .
\end{equation}
For the ensemble the smallest positive probability is 0.0941 and $\theta^{\star}=0.0940$. At $\theta^{\star}$ we report precision, false-alarm rate (false positives over negatives), F1 and the Matthews correlation coefficient~\citep{chicco2020advantages}. We also report threshold-free ROC-AUC and PR-AUC (average precision), reading PR-AUC against the base rate~\citep{saito2015precision}, and the maximum-F1 point for reference. Since $\theta^{\star}$ is fixed on the validation set, the 100\% recall is an in-sample property of the threshold, not an out-of-sample guarantee.

\paragraph{Same-date control.}
\label{sec:sdc-protocol}
To test whether the model separates clusters for reasons other than the calendar, we compute the AUC among clusters on the same date. Within one date every cluster shares the same calendar features, so the calendar contributes no information and the within-date AUC measures spatial discrimination only. In the labels used for the main model, no negative exists on any October or November 2025 date, so the control requires its own design. We built a separate label set in which, on the 37 dates with positive clusters, the other 75 clusters are labeled negative (3,515 samples: 380 positive, 3,135 negative), and trained the same architecture with five-fold cross-validation over clusters and three seeds (15 models). The metric is the mean within-date AUC of out-of-fold predictions over 17 November dates, with a cluster bootstrap (95 clusters, 2,000 resamples). The pass criterion (lower bound above 0.5) was fixed before measurement. This control tests the architecture and training procedure, not the deployed checkpoints.

\paragraph{Pre-registration.} For the robustness experiments in Sections~\ref{sec:expA}--\ref{sec:din}, the acceptance criterion was written down before the measurement and not changed afterwards: a degraded-input condition is ``usable'' only if detection stays at 100\% and AUC drops by less than 0.02. When a criterion later proved to be badly posed, we recorded it as a criterion error rather than reinterpreting the result.

\paragraph{Baselines.} A separate benchmark compared \stmmt{} with three baselines trained on the same samples, normalization, loss (focal loss with class weights 0.5/1.5 and confidence weighting) and optimizer settings (Adam, learning rate $10^{-4}$ with cosine annealing, gradient clipping 1.0, batch 16, at most 30 epochs, patience 8): a two-layer LSTM~\citep{hochreiter1997long} with hidden size 256 applied to the spatial mean of each channel, giving a $14\times32$ sequence; a VGG16-configuration network~\citep{simonyan2015very}, without pretrained weights, applied to the temporal mean of the input; and XGBoost~\citep{chen2016xgboost} with 400 trees of depth 6 on 160 hand-crafted features (mean, standard deviation, trend, last value and spatial standard deviation for each of the 32 channels). Each baseline was trained with seeds 42, 1 and 7, and each model's threshold was set by the same recall-100\% rule on its own predictions. No model received a hyperparameter search.

\section{Results}
\label{sec:results}

Except for the same-date control (Section~\ref{sec:sdc}), which comes from a separate cross-validated training, all results in this section are for the six-seed ensemble of \texttt{xattn\_v3\_focal\_7day} on the validation set of 2,792 samples (8 clusters, 2024--2025; 176 positives, base rate 6.30\%). The same samples were used to select checkpoints and to set the operating threshold, so the figures describe fit to the validation set rather than performance on unseen data. Measurement dates are listed in Appendix~\ref{app:provenance}.

\subsection{Overall accuracy}
Table~\ref{tab:main} summarizes accuracy. The ensemble ranks positives above negatives almost perfectly (ROC-AUC 0.9940), and its PR-AUC of 0.9097 is 14.4 times the base rate. At the operating threshold $\theta^\star=0.0940$ every reported positive is detected, 92 of 2,616 negatives are flagged (3.5\%), and precision is 65.7\%. Choosing the maximum-F1 threshold instead would raise F1 from 0.7928 to 0.8721 but miss 26 of the 176 positives (recall 85.2\%), which is why we do not use it operationally.

\begin{table}[t]
\centering
\begin{threeparttable}
\caption{Validation accuracy of the six-seed \stmmt{} ensemble (left) and the confusion matrix at the operating threshold $\theta^\star=0.0940$ (right).}
\label{tab:main}
\small
\begin{tabular}{@{}lr@{\hspace{30pt}}lrr@{}}
\toprule
Metric & Value & & Predicted + & Predicted $-$ \\
\midrule
ROC-AUC & 0.9940 & Reported positive & 176 & 0 \\
PR-AUC (average precision) & 0.9097 & Negative (2024) & 92 & 1,476 \\
Recall at $\theta^\star$ & 100.0\% & Negative (2025) & 0 & 1,048 \\
Precision at $\theta^\star$ & 65.7\% & Negative, total & 92 & 2,524 \\
False-alarm rate at $\theta^\star$\tnote{a} & 3.5\% & & & \\
F1 at $\theta^\star$ & 0.7928 & & & \\
MCC at $\theta^\star$\tnote{b} & 0.7960 & & & \\
Maximum F1 (recall 85.2\%) & 0.8721 & & & \\
\bottomrule
\end{tabular}
\begin{tablenotes}\footnotesize
\item[a] Upper bound: negatives are unreported days, not confirmed healthy (Section~\ref{sec:labels-used}).
\item[b] Computed from the confusion matrix.
\end{tablenotes}
\end{threeparttable}
\end{table}

All false alarms occur in 2024, the year without reported damage, and none among the 1,048 negatives of 2025. The most likely explanation is the season: every 2025 negative has a target date between January and May, whereas the 2024 negatives cover the whole year, including the autumn months in which the model raises its probabilities. A second possibility is that some 2024 alarms reflect real but unreported stress, which the positive--unlabeled structure cannot rule out. With two years of data we cannot separate these explanations.

\subsection{Alert grades}
\begin{figure}[t]
\centering
\includegraphics[width=0.72\linewidth]{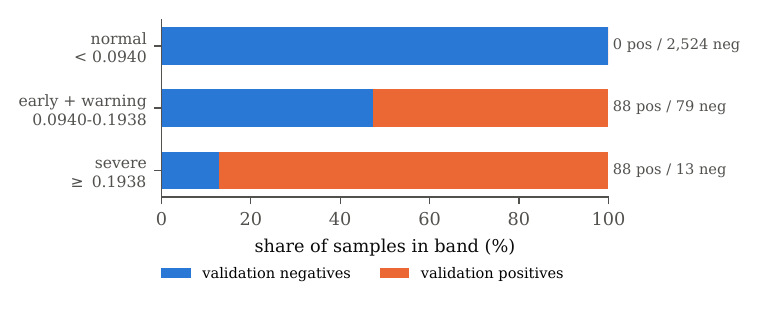}
\caption{Validation samples by ensemble probability band. Boundaries are the recall-100\% threshold (0.0940) and the median of positive probabilities (0.1938); the early-warning and warning grades share the middle band (split at 0.1120).}
\label{fig:bands}
\end{figure}

The service shows four grades. Their boundaries are fixed by the distribution of validation positives rather than chosen by hand: $\theta^\star=0.0940$ separates normal from early warning, the lower quartile of positive probabilities (0.1120) separates early warning from warning, and the median (0.1938) separates warning from severe. (One internal report describes 0.1120 and 0.1938 as the median and upper quartile; the band counts below, with 88 of the 176 positives at or above 0.1938, confirm that 0.1938 is the median.) Figure~\ref{fig:bands} shows how the validation samples fall into the three resulting bands. No positive falls in the normal band, which holds 2,524 negatives. The severe band holds 88 positives and 13 negatives, and the middle band holds 88 positives and 79 negatives. The early band is only 0.018 wide, and in October 2025 a third of the forecasts fall into it; small changes of the threshold would therefore move many farms between grades, which is why the boundaries are pre-registered rather than tuned.

\subsection{Same-date control}
\label{sec:sdc}
The calendar shortcut in Section~\ref{sec:labels-used} means that a high AUC alone could reflect the season rather than the ocean. Table~\ref{tab:sdc} reports the within-date AUC of the separate cross-validated training described in Section~\ref{sec:protocol}: on November 2025 dates, across 95 clusters, out-of-fold predictions rank the clusters reported as bleached above the others with AUC 0.976--0.986, and all bootstrap lower bounds exceed 0.94, well above the pre-registered criterion of 0.5. With the calendar held fixed, the architecture trained under this design still separates clusters.

\begin{table}[t]
\centering
\begin{threeparttable}
\caption{Same-date control: within-date AUC among 95 clusters on November 2025 dates (out-of-fold predictions of the cross-validated control training), where calendar features are identical for all clusters.}
\label{tab:sdc}
\small
\begin{tabular}{@{}ccc@{}}
\toprule
Replicate\tnote{a} & Within-date AUC & Bootstrap lower bound \\
\midrule
1 & 0.9778 & 0.9466 \\
2 & 0.9763 & 0.9429 \\
3 & 0.9857 & 0.9621 \\
\bottomrule
\end{tabular}
\begin{tablenotes}\footnotesize
\item[a] Our records label the three rows differently (seeds 1--3; a design that fixed seeds 42, 1 and 7; or three November date groups). The values are the same in all records.
\end{tablenotes}
\end{threeparttable}
\end{table}

Three limits apply. First, the control was run on a separately trained set of models, not on the deployed ensemble, whose own calendar independence has not been measured; the deployed forecast pack still assigns the early-warning grade to about 40\% of farms in September of both years (36.8\% in 2024 and 44.3\% in 2025), a residual seasonal baseline. Second, the control removes the calendar but not every confounder: on a fixed date, clusters still differ in persistent characteristics such as location and typical nutrient levels, and the 20 positive clusters may share such traits. Third, the clusters labeled negative on November dates are unreported, not confirmed healthy. The control therefore shows spatial discrimination, not that the model has learned the causal pathway to bleaching. It was introduced after we found that an earlier forecast pack behaved as a calendar switch (Section~\ref{sec:pack}).

\subsection{Comparison with baselines}
Table~\ref{tab:bench} reports the benchmark run of 7 October 2026 on the same 2,792 validation samples (training: 27,784 samples, 204 positives, 87 clusters). Each baseline was trained with three seeds; its row is the mean of the per-seed values, each computed at that seed's own recall-100\% threshold. \stmmt{} is shown both as the mean of its six single-seed checkpoints and as the six-seed ensemble.

\begin{table}[t]
\centering
\begin{threeparttable}
\caption{Benchmark at each model's own recall-100\% threshold. Single-model rows are means over seeds; the ensemble row is computed from averaged probabilities.}
\label{tab:bench}
\small
\begin{tabular}{@{}lrccccc@{}}
\toprule
Model & Parameters & ROC-AUC & PR-AUC & Precision & False alarm & MCC \\
\midrule
XGBoost (3 seeds) & --\tnote{a} & 0.9150 & 0.6288 & 14.62\% & 39.67\% & 0.2970 \\
VGG16 (3 seeds)\tnote{b} & 39,913,602 & 0.9810 & 0.7704 & 41.49\% & 36.44\% & 0.7505 \\
LSTM (3 seeds) & 823,810 & 0.9953 & 0.9487 & 60.15\% & 4.74\% & 0.7550 \\
\stmmt{} (6 single seeds) & 9,218,179 & 0.9907 & 0.8583 & 48.71\% & 7.67\% & 0.6681 \\
\stmmt{} (6-seed ensemble)\tnote{c} & $6\times$9,218,179 & 0.9940 & 0.9096 & 65.67\% & 3.52\% & 0.7960 \\
\bottomrule
\end{tabular}
\begin{tablenotes}\footnotesize
\item[a] About 1,000 tree leaves per model.
\item[b] One seed assigned a near-zero probability to one positive, so its recall-100\% threshold fell to 0 and every sample was flagged (MCC undefined); the MCC shown averages the other two seeds.
\item[c] PR-AUC 0.9096 here against 0.9097 in Table~\ref{tab:main}, from a separate evaluation of the same checkpoints.
\end{tablenotes}
\end{threeparttable}
\end{table}

\begin{figure}[t]
\centering
\includegraphics[width=0.6\linewidth]{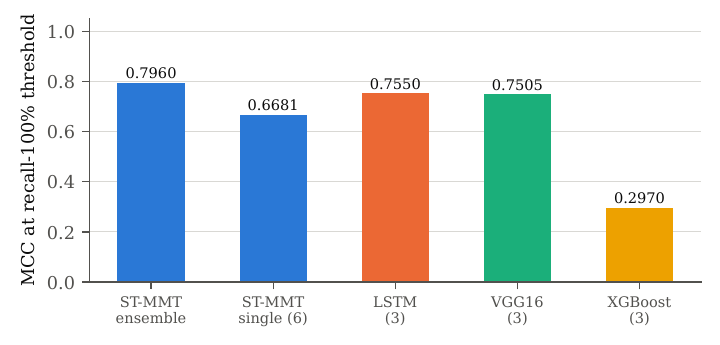}
\caption{Matthews correlation coefficient at each model's recall-100\% threshold (benchmark of 7 October 2026). Single-model bars are means over seeds; the VGG16 bar excludes one seed whose threshold collapsed to zero.}
\label{fig:bench}
\end{figure}

Figure~\ref{fig:bench} summarizes the operating-point comparison. The result does not favour \stmmt{} as a single model. Model for model, the LSTM on spatially averaged channels is better on every metric: higher ranking accuracy (ROC-AUC 0.9953 against 0.9907, PR-AUC 0.9487 against 0.8583) and, at the operating point, higher precision (60.15\% against 48.71\%), fewer false alarms (4.74\% against 7.67\%) and a higher MCC (0.7550 against 0.6681). The \stmmt{} ensemble has the best operating-point figures in the table (precision 65.67\%, false alarms 3.52\%, MCC 0.7960), but it is compared with the average of single LSTMs, which is not a like-for-like comparison, and the best single LSTM (seed 1: precision 73.6\%, false alarms 2.4\%, MCC 0.848) exceeds it. The variation between seeds is large for both families: single-seed \stmmt{} false-alarm rates range from 3.7\% to 10.8\%, and single-seed LSTM rates from 2.4\% to 6.6\%.

One feature of the LSTM runs calls for caution before reading the comparison as evidence that space does not matter. In all three seeds the training loss fell to about zero within the first two epochs, and two seeds reached their best validation score in the first epoch. The LSTM receives the spatial means of all channels, including the month encodings, and the labels can be separated perfectly by year and month (Section~\ref{sec:labels-used}). A model that learns the season will look excellent on this validation set. We have not tested whether the LSTM relies on the calendar. A pre-registered follow-up trains the baselines as six-seed ensembles, compares ensembles with bootstrap intervals, and submits the baselines to the same-date control of Section~\ref{sec:sdc}; it will be reported whatever its outcome. Until then, the defensible reading is that the current labels do not demonstrate an advantage of the spatiotemporal transformer over a temporal model of cluster averages, and that the transformer's practical additions are the within-cluster probability map and an architecture in which modalities can interact explicitly.

\subsection{Robustness to missing channels}
\label{sec:expA}
In operation some channels arrive late or not at all. Experiment A simulated this for nine channels that depend on separate collection steps (currents $u$, $v$ and speed; SST anomaly; SST 7- and 30-day means; chlorophyll 7-day mean; NIR index and its daily change). Two failure modes were compared: values frozen in time (flattened) and values set to zero (Table~\ref{tab:expA}).

\begin{table}[t]
\centering
\caption{Experiment A: effect of degrading nine channels (ch10--13, 24, 27--30). Pre-registered criterion: detection 100\% and AUC drop $<0.02$.}
\label{tab:expA}
\small
\begin{tabular}{@{}lccccc@{}}
\toprule
Condition & ROC-AUC & Detection & False alarm & False alarm 2024 & Verdict \\
\midrule
Original & 0.9940 & 100.0\% & 3.5\% & 5.9\% & reference \\
Nine channels flattened in time & 0.9943 & 100.0\% & 3.9\% & 6.4\% & usable \\
Nine channels set to zero & 0.9778 & 88.6\% & 3.1\% & 5.2\% & not usable \\
\bottomrule
\end{tabular}
\end{table}

Frozen values are harmless, but zeros are not: 20 of the 176 positives are missed. The lower false-alarm rate in the zero condition is not an improvement; the model simply outputs lower probabilities overall. This matters directly for operation, because the 2026 inference patches checked on 7 October 2026 had exactly ten channels at zero (the nine above plus mixed-layer depth). The cause was a build script that skipped several processing steps, not missing data, and the failure would have recurred on any machine (Section~\ref{sec:limits}). A follow-up recorded in our nutrient-estimation log reports AUC 0.9768 and detection 88.6\% with all ten channels at zero, and AUC 0.9850 with 100\% detection after restoring the moving averages, currents and mixed-layer depth; our training report still lists these conditions as unmeasured, so we cite them only as preliminary.

\subsection{Inference port and forecast pack}
\label{sec:pack}
The service does not run the network. A stand-alone inference module produces probabilities that are written into a forecast pack, and the web service reads the pack. We compared the stand-alone module with the training code on all 2,792 validation samples: the maximum absolute difference was $2.3\times10^{-6}$ and no sample differed by more than $10^{-4}$.

Auditing the forecast pack that the service used before regeneration, which carried the same model name but was produced by a different generator, revealed a calendar switch (Table~\ref{tab:pack}). Within a month, almost every farm received the same grade: 100\% severe in September 2024 and 92.5\% in October 2024, although 2024 was a year without reported damage, and on 1 November 2025 the risk scores of eight clusters spanned a range of only 0.013. The regenerated pack from the ensemble has no severe grades in any month of 2024, concentrates severe grades in October--December 2025 (15.0\%, 43.8\% and 24.7\% of farms), and spans a range of 0.509 on the same date. Regeneration also fixed three defects in the original generator: dates were collapsed because only the last row per tile was kept, the ``severe'' field was a fixed multiple ($1.2\times$) of the mean probability instead of a model output, and grade boundaries were arbitrary cut-offs on ten times the probability. The pack now covers 469 dates (21 January 2024 to 31 December 2025) and 1,109 farms; the grade distribution is 82.9\% normal, 9.9\% early, 1.6\% warning and 5.6\% severe, and 136 farms are severe on 14 October 2025.

\begin{table}[t]
\centering
\caption{Forecast pack before and after regeneration with the ensemble.}
\label{tab:pack}
\small
\begin{tabular}{@{}lcc@{}}
\toprule
Check & Previous pack & Regenerated pack \\
\midrule
Share severe, September 2024 & 100\% & 0.0\% \\
Share severe, October 2024 & 92.5\% & 0.0\% \\
Share severe, November 2025 & 10.9\% & 43.8\% \\
Share of ``early'' grade & 0.0\% (all 17 months) & 9.9\% \\
Range across clusters, 1 Nov 2025 & 0.013 & 0.509 \\
\bottomrule
\end{tabular}
\end{table}

One check from the regenerated pack remains open: 24.7\% of farms are graded severe in December 2025, and we have not yet compared this with damage records for that month.

\section{The nutrient gap in operation}
\label{sec:din}

DIN is the nutrient most closely associated with bleaching in the literature~\citep{hori2008discoloration,nishikawa2011eucampia}, and it is the input that is hardest to obtain in real time. DIN is not measured directly: it is the sum of three separately analysed species,
\begin{equation}
  \mathrm{DIN} = [\mathrm{NO_3^-}\text{-N}] + [\mathrm{NO_2^-}\text{-N}] + [\mathrm{NH_4^+}\text{-N}],
\end{equation}
so a DIN value exists only for days on which water was sampled and analysed. This section quantifies the gap and evaluates the obvious ways of bridging it.

\subsection{How old is the newest nutrient observation?}
Table~\ref{tab:gap} lists, for each source we ingest, the most recent nutrient observation available on 7 October 2026. KOEM surveys are quarterly (February, May, August and November). The late-August surveys of 2024 and 2025 are in our database; the 2026 August survey was not available, and manual queries for four sea areas and two periods returned no records and no errors, which points to delayed publication rather than a failure of our collector. The freshest usable nutrient value was therefore 132 days old. The NIFS station networks had no 2026 nutrient records, and the NIFS real-time stations report temperature only. CMEMS daily biogeochemistry is current but is a model product without ammonium, and it was not used to train the model.

\begin{table}[t]
\centering
\caption{Most recent nutrient observation per source on 7 October 2026.}
\label{tab:gap}
\small
\begin{tabular}{@{}llrl@{}}
\toprule
Source & Latest nutrient value & Age (days) & Note \\
\midrule
KOEM monitoring network & 28 May 2026 & 132 & August survey not yet published \\
NIFS serial oceanographic stations & 16 Dec 2025 & 295 & no 2026 nutrient records \\
NIFS fishing-ground stations & 5 Nov 2025 & 336 & sparse responses \\
NIFS real-time stations & -- & -- & temperature only \\
CMEMS biogeochemistry (model) & 5 Oct 2026 & 2 & daily; not used in training \\
\bottomrule
\end{tabular}
\end{table}

\subsection{Cost of stale nutrients}
\label{sec:expB}
\begin{figure}[t]
\centering
\includegraphics[width=0.92\linewidth]{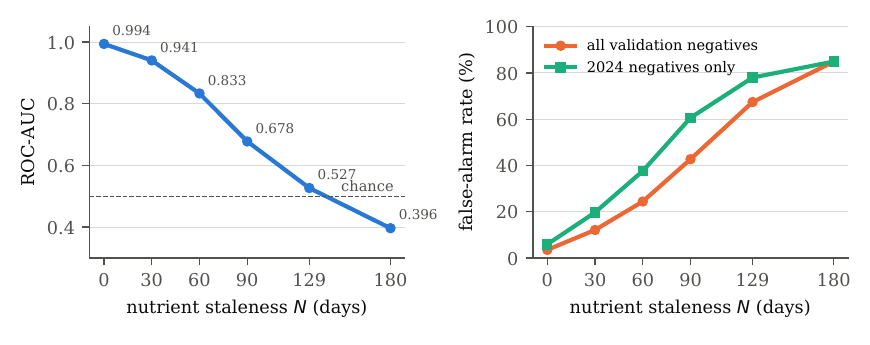}
\caption{Experiment B: ROC-AUC (left) and false-alarm rate (right) when the eleven KOEM channels are frozen $N$ days before the input window. At 129 days, close to the 132-day age of the freshest nutrient value on 7 October 2026, discrimination is close to chance.}
\label{fig:stale}
\end{figure}

Experiment B froze the eleven KOEM channels at their values $N$ days before each input window and re-evaluated the ensemble (Figure~\ref{fig:stale} and Table~\ref{tab:expB}). Two reading rules apply. First, the training data already contain a median natural staleness of 17 days, so the effective age is about $N+17$. Second, the number of evaluable samples falls from 2,537 to 820 as $N$ grows, and the composition changes; detection rates at 129 and 180 days are therefore not interpretable, and only AUC and false-alarm rate are reported for those rows.

\begin{table}[t]
\centering
\caption{Experiment B in detail. Detection rates for 129 and 180 days are omitted because the evaluable sample changes composition.}
\label{tab:expB}
\small
\begin{tabular}{@{}rrcccc@{}}
\toprule
$N$ (days) & Samples & ROC-AUC & Detection & False alarm & False alarm 2024 \\
\midrule
0 & 2,792 & 0.9940 & 100.0\% & 3.5\% & 5.9\% \\
30 & 2,537 & 0.9405 & 97.7\% & 12.1\% & 19.7\% \\
60 & 2,087 & 0.8333 & 67.6\% & 24.4\% & 37.6\% \\
90 & 1,637 & 0.6778 & 87.5\% & 42.7\% & 60.6\% \\
129 & 1,052 & 0.5267 & -- & 67.4\% & 78.0\% \\
180 & 820 & 0.3964 & -- & 84.9\% & 84.9\% \\
\bottomrule
\end{tabular}
\end{table}

Under the pre-registered criterion not even the 30-day condition is acceptable: AUC drops by 0.0535 and detection falls to 97.7\%. At 129 days the AUC (0.5267) is indistinguishable from chance, and at 180 days it is below 0.5. Two practical conclusions follow. The 60-day freshness limit used by our output gate (Section~\ref{sec:gate}) has no empirical basis, because every measured condition fails; a defensible limit requires measurements at 7, 14 and 21 days, which we have not yet made. And during the months when the newest survey is three to four months old, the model should not issue nutrient-dependent alerts at all.

\subsection{Decomposing DIN for buoy measurements}
A buoy with nitrate and ammonium sensors, but no nitrite sensor, is planned for a partner farm in Goheung. To estimate the bias of $\mathrm{NO_3}+\mathrm{NH_4}$ as a DIN measurement, we analysed 24,513 NIFS fishing-ground samples in which all three species were reported (Table~\ref{tab:no2}). The sum of the three species matches the reported DIN (median relative error 0.0000; 5.2\% of samples differ by more than 10\%), so the nitrite share can be trusted. Nitrite contributes a median 5.26\% of DIN (mean 8.07\%, 95th percentile 23.84\%). If nitrite is a fraction $f$ of DIN, then $\mathrm{DIN} = (\mathrm{NO_3}+\mathrm{NH_4})/(1-f)$. Once the buoy is deployed, we plan to correct its readings with the median factor $1/(1-0.0526)=1.055$ and to use the 95th-percentile factor $1/(1-0.2384)=1.313$ for decisions near alert thresholds, where under-estimation would suppress alerts. (Our internal report rounds the conservative factor to 1.24, i.e.\ $1+0.2384$, which under-corrects.) For scale, KOEM DIN values ($n=23{,}597$) have mean 138.41, median 76.57 and 90th percentile 314.02\,$\mu$g/L.

\begin{table}[t]
\centering
\caption{Share of nitrite in DIN (NIFS fishing-ground samples, $n=24{,}513$).}
\label{tab:no2}
\small
\begin{tabular}{@{}lr@{}}
\toprule
Statistic & Value \\
\midrule
NO$_2$ share, median / mean / 95th percentile & 5.26\% / 8.07\% / 23.84\% \\
Median relative error, sum of species vs.\ reported DIN & 0.0000 \\
Samples with relative error $>10\%$ & 1,278 (5.2\%) \\
Correction factor $1/(1-f)$ for NO$_3$+NH$_4$ (median / 95th percentile) & 1.055 / 1.313 \\
\bottomrule
\end{tabular}
\end{table}

\subsection{Persistence versus climatology}
Forward filling a 132-day-old May value into October assumes that DIN has not changed, an assumption that had never been tested. We compared it with a station climatology using pairs of consecutive KOEM observations at 425 stations from 2011 to 2026 (22,736 pairs). For a pair $(t_k, t_{k+1})$ at station $s$, persistence predicts $D(s,t_{k+1}) = D(s,t_k)$, and climatology predicts
\begin{equation}
  C(s,d) = \frac{1}{|\mathcal{Y}|}\sum_{y\in\mathcal{Y}} \operatorname{mean}\big\{D(s,t): \mathrm{year}(t)=y,\ |\mathrm{doy}(t)-d|_{\circ}\le 15\big\},
\end{equation}
where $d$ is the day of year of $t_{k+1}$, $|\cdot|_{\circ}$ is the circular day-of-year distance, and $\mathcal{Y}$ excludes the evaluation year (leave-one-year-out) and must contain at least three years. The pre-registered criterion was at least 100 pairs and an MAE improvement of at least 20\%.

\begin{figure}[t]
\centering
\includegraphics[width=0.85\linewidth]{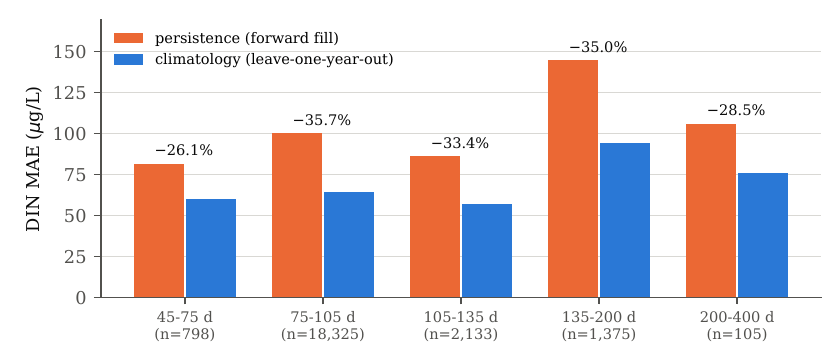}
\caption{DIN prediction error of persistence (forward fill) and leave-one-year-out climatology for consecutive KOEM observations, grouped by the gap between observations. Labels give the relative reduction in MAE achieved by climatology.}
\label{fig:clim}
\end{figure}

Climatology won in every gap bucket, by 26.1\% to 35.7\% (Figure~\ref{fig:clim}); at the current gap of 105--135 days the MAE fell from 86.03 to 57.26\,$\mu$g/L (33.4\%). The persistence error at that gap is larger than the median DIN value itself (76.57\,$\mu$g/L), which is consistent with the near-chance AUC of Experiment B. This is, however, an error in DIN, not in bleaching detection. Climatology varies less than forward-filled values and may lie outside the distribution the model saw in training, and it does not change from day to day. We have not yet measured detection accuracy with climatological inputs, and we have not changed the operational input.

\subsection{A rejected proxy: nutrient drift from daily variables}
\label{sec:nde}
The Nutrient Drift Estimator (NDE) tried to fill the gap with change rather than level: find a channel that is observed daily and whose changes track changes in DIN, and propagate that change,
\begin{equation}
  \widehat{D}(t) = D(t_k) + \beta\,\big[P(t)-P(t_k)\big],
\end{equation}
where $t_k$ is the last survey and $P$ is the proxy. We first separated candidate channels by whether they are measured in the same survey as DIN; same-survey channels such as nitrate, phosphate and silicate track DIN closely (change correlations 0.71--0.95) but go stale together with it, so they cannot fill the gap. Among daily channels, the pre-registered criterion required an absolute change correlation of at least 0.30 and a top-three rank in both 2024 and 2025.

\begin{table}[t]
\centering
\caption{Change correlation with DIN for the leading candidates. The daily candidates that ranked highest in 2024 are seasonal variables whose correlation collapses in 2025.}
\label{tab:nde}
\small
\begin{tabular}{@{}llrr@{}}
\toprule
Group & Channel & 2024 & 2025 \\
\midrule
\multirow{3}{*}{Same survey (cannot fill the gap)} & NO$_3$ (ch21) & 0.945 & 0.894 \\
 & DIP (ch2) & 0.800 & 0.809 \\
 & SiO$_2$ (ch3) & 0.710 & 0.842 \\
\midrule
\multirow{3}{*}{Daily (candidates)} & Air temperature (ch19) & $-0.501$ & $-0.089$ \\
 & Month cosine (ch26) & 0.461 & 0.168 \\
 & Month sine (ch25) & 0.345 & 0.046 \\
\bottomrule
\end{tabular}
\end{table}

The three best daily candidates in 2024 were air temperature and the two month encodings (Table~\ref{tab:nde}), that is, the season: DIN is high in winter and low in summer. In 2025, the year with bleaching, their correlations collapsed to $-0.089$, 0.168 and 0.046. Without the requirement that a proxy rank highly in both years, we would have adopted air temperature on the strength of 2024 alone, and it would have failed in the year that mattered. Two follow-up variants were also rejected under their pre-registered criteria, according to our nutrient-estimation log: summary statistics over the gap interval (for every one of 103 combinations, at least one of the two directions increased the error; the best combination, the interval maximum of the wind-direction cosine, improved one direction by 0.7\% and worsened the other by 0.8\%), and ridge regression of log DIN on same-day ocean-model and calendar fields, which was worse than climatology by 244.6\% (2024$\rightarrow$2025) and 55.0\% (2025$\rightarrow$2024). Our training report of the same date still lists the interval-summary measurement as pending, so these two results have not been cross-checked. One observation from this search is worth keeping: level and change correlations can differ sharply (ammonium: 0.126 for level and 0.596 for change; dissolved oxygen: 0.019 and $-0.585$), so external sources should not be dismissed on level correlation alone.

\subsection{The output gate}
\label{sec:gate}
Forecasts pass through a three-stage gate before they are shown, and each stage fails closed, meaning that when the system cannot decide, it does not alert (Table~\ref{tab:gate}).

\begin{table}[t]
\centering
\caption{Harness Gate applied to every forecast before display.}
\label{tab:gate}
\small
\begin{tabularx}{\linewidth}{@{}lXl@{}}
\toprule
Stage & Rule & Default \\
\midrule
1. Freshness & block if the newest nutrient observation is older than the limit & 60 days \\
2. Data quality & require declared valid channels; otherwise mark the forecast ``unknown'' & -- \\
3. Persistence & raise a grade after consecutive exceedances, lower it after consecutive clear days & 2 / 5 / 7 days \\
\bottomrule
\end{tabularx}
\end{table}

With the newest nutrient value 132 days old, stage 1 blocks every farm. Given Experiment B, this is the correct behaviour: at that age the false-alarm rate is 43--67\%, and repeated false alarms would teach users to ignore the system. The service response carries the age of the nutrient data, and the pack declares its own alert threshold (0.0940), which overrides the gate's default persistence threshold of 0.30. The service also refuses to serve a pack whose model name does not match the deployed model (HTTP 503) rather than fall back silently to an older model.

\section{Development history and lessons}
\label{sec:history}

The deployed model is the result of more than twenty training configurations. Their metrics are not comparable with the results above: early runs used a single 2021 tile (Wando) and later two tiles (Wando and Mokpo), counted pixels rather than samples, and, until a code review found the error, predicted the next day rather than seven days ahead. We summarize the lessons rather than the numbers; Appendix~\ref{app:history} lists the key configurations with these caveats.

\paragraph{Diagnose suspiciously good results.} A model trained on the Mokpo tile alone reached ROC-AUC and PR-AUC of exactly 1.0000. We treated this as an artifact rather than a success; it persisted after removing a suspect channel and was not used. The same attitude found the N:P ratio bug (values up to $2.65\times10^6$ that dominated the saliency map) and a suspended-solids channel whose apparent importance came from having only four to seven distinct values per year.

\paragraph{The 2021 South Jeolla experiments.} Development began with a single 2021 tile at Wando, where an official record placed a bleaching event from 6 September to 26 October 2021 (51 days); with the remaining 314 days labeled normal, one tile-year gave 345 sliding-window samples (300 for training, 45 for validation). Mokpo was added as a second tile, and Jindo and then Goheung (with a December 2016 event) as third and fourth. Performance fell as tiles were added: ROC-AUC 0.9071, 0.8378 and 0.7790 and PR-AUC 0.8522, 0.8091 and 0.7212 for two, three and four tiles (validation sets of 90, 135 and 180 samples). Tracing the drop exposed data problems rather than a modelling limit. The tile centres for Jindo and Goheung lay 19.65\,km and 23.78\,km from the actual farms, outside a tile that is 15.36\,km wide, against 4.53\,km for Wando and 6.15\,km for Mokpo; the Goheung tile had only 28 of 32 channels, with no satellite data before 2020, most solar-radiation values missing and no ocean-model currents for January--April 2016; and the Jindo tile lacked mixed-layer depth. The project subsequently moved to the national register of licensed farm polygons, cluster tiles built around those farms, and the 2024--2025 two-year design used in the deployed model. The 2021 numbers were counted in pixels and, as noted above, predicted the next day, so they are not comparable with Section~\ref{sec:results}; Appendix~\ref{app:history} lists them.

\paragraph{Cross-attention alone was not enough.} The first cross-attention model, with the three groups encoded separately and fused through the Group A hub, improved ROC-AUC over its self-attention predecessor (0.9191 against 0.8716 on the two-tile, next-day task) but lowered F1 at the recall-100\% threshold (0.8013 against 0.8953) and precision (75.36\% against 84.75\%). Adding a fourth nutrient group did not help (F1 0.7786), and neither did fully pairwise bidirectional cross-attention (0.7343), forcing the residual gate to start at 1.0 (0.8507) or hard-negative mining (0.8059). Hard-negative mining on the four-group model reached F1 0.8985 but lowered its probability for an external event to 8.2\%, so it was discarded; the residual design with focal loss was the first configuration that exceeded the self-attention model at the operating point and was retained (F1 0.9271, precision 88.74\%). In the residual variant trained before focal loss was introduced, the learned gate ended at $\alpha=0.105$ from an initial 0.1, indicating that the cross-attention path contributed little in that setting. We have not recorded the gate value of the deployed checkpoints, and we have not run an ablation that removes the group path from the final configuration. The contribution of cross-attention to the results in Section~\ref{sec:results} is therefore not isolated, and we regard measuring it as the most important open experiment for the architecture.

\paragraph{The seasonal confound and the retraction of an external validation.} An external check on the Seocheon coast in 2025 initially appeared to catch a large event on 14 October. An annual inference curve later showed that the model of that period assigned probabilities of 0.70--0.80 in July and August but 0.11 in November and December, and that its output correlated most strongly with the 30-day SST mean (0.843), the month sine ($-0.812$) and air temperature (0.802). It had learned the season. We withdrew the external validation and retrained with a design that contrasts the same months of a year without reported damage (2024) with the event year (2025). Table~\ref{tab:ch20} summarizes the retraining on eight external clusters. Restoring channels improved AUC but left a strong summer response (November/summer ratio 1.41 and 132 alarm days in 2024); excluding clusters without observations and restoring SST anomaly and mixed-layer depth gave a ratio of 9.42 and no alarm days in 2024 at a threshold of 0.50 (30 days at 0.30); replacing KOEM nutrients with CMEMS model nutrients lowered AUC to 0.9355. Since then an annual inference curve has been a required check before any model is adopted, because validation metrics cannot reveal seasonal dependence when no labels exist for the summer months.

\begin{table}[t]
\centering
\caption{Retraining against the seasonal confound (8 external clusters, three seeds, mean $\pm$ s.d.). The ratio is the mean probability in November 2025 divided by that in June--August; alarm days count 2024 days above an alert threshold, which was 0.30 for the second row and 0.50 for the third (at 0.30 the third configuration has 30 alarm days).}
\label{tab:ch20}
\footnotesize
\setlength{\tabcolsep}{4pt}
\begin{tabular}{@{}lcccc@{}}
\toprule
Configuration & ROC-AUC & F1 (recall 100\%) & Nov./summer ratio & Alarm days 2024 \\
\midrule
Original patches (22 valid channels) & 0.9825 & 0.7756 & -- & -- \\
Moving averages and currents restored (29) & $0.9922\pm0.0053$ & $0.8656\pm0.0806$ & 1.41 & 132 \\
Empty clusters excluded, SST anomaly and MLD & $0.9820\pm0.0067$ & $0.7647\pm0.0560$ & 9.42 & 0 \\
KOEM nutrients replaced by CMEMS & $0.9355\pm0.0092$ & $0.6929\pm0.0136$ & -- & -- \\
\bottomrule
\end{tabular}
\end{table}

\paragraph{Separate ranking metrics from operating metrics.} F1 at the recall-100\% threshold depends on the single lowest-scoring positive and therefore varies widely between seeds (0.8087 to 0.9578 for one configuration). We use AUC to compare models and report precision, false alarms and F1 at the operating point to describe how a model will behave in service.

\section{Limitations}
\label{sec:limits}
\begin{itemize}
  \item \textbf{Two years and a small validation set.} Training contrasts one year without reported bleaching with one year with bleaching. The validation set covers 8 clusters and was also used to select checkpoints and the threshold; the false-alarm rate of 3.5\% is measured on those clusters only and need not hold for all 105. There is no separate test set.
  \item \textbf{Report-based, positive--unlabeled labels.} All positives fall in October and November 2025, so the year and month alone separate the labels. The same-date control was run on a separately trained set of models; the calendar independence of the deployed ensemble is not measured, and its pack still shows a raised early-warning share in September of both years. False-alarm rates are upper bounds.
  \item \textbf{Non-causal backward fill.} 22.3\% of the training samples (January--February targets) contain values back-filled from the first survey of the year. The operational path uses forward fill only, and retraining without the backward pass is pending.
  \item \textbf{Cross-attention not isolated.} The contribution of the cross-attention path in the deployed configuration has not been measured by ablation, and the BLC soft labels were not the training target of the evaluated model.
  \item \textbf{Baselines.} A single LSTM on cluster-average series outperforms a single \stmmt{} on every benchmark metric, and only the \stmmt{} ensemble leads at the operating point, against single-seed baselines. Ensemble-to-ensemble comparison and a same-date control for the baselines are pre-registered but not yet run.
  \item \textbf{Nutrient staleness.} Skill decays steeply with the age of the nutrient survey (Section~\ref{sec:expB}), and the 2026 August survey had not been published at the time of writing. The gate's 60-day limit is not yet grounded in measurement.
  \item \textbf{2026 inputs.} The 2026 inference patches checked on 7 October had ten channels at zero because the build skipped several processing steps. The moving averages, currents and mixed-layer depth can be restored from stored data; the SST anomaly needs its multi-year baseline, whose location must still be confirmed; the satellite NIR channels require new collection. Preliminary measurements of the restored conditions (Section~\ref{sec:expA}) have not been cross-checked.
  \item \textbf{Unmeasured items.} Inference speed on CPU or edge hardware, the accuracy of a distilled model, detection accuracy with climatological nutrient inputs, freshness limits shorter than 30 days, and the December 2025 severe grades against damage records.
\end{itemize}

\section{Conclusion and future work}
\label{sec:conclusion}
We described \stmmt{}, a residual spatiotemporal transformer with hub-style multi-head cross-attention that forecasts laver bleaching seven days ahead from 32 aligned channels, together with the data pipeline (\otc{}), the labeling procedure (BLC) and the evaluation protocol around it. On the validation set the six-seed ensemble detects every reported positive at a false-alarm rate of at most 3.5\%, and a same-date control on a separately trained set of models shows that the architecture can discriminate between clusters when the calendar is held fixed. The same evaluation shows where the system is fragile: performance collapses when the quarterly nutrient survey is a few months old, and zeros in place of missing channels cost 20 of the 176 detections.

Three lines of work follow directly. First, \emph{measure what is unmeasured}: an ablation of the cross-attention path, operating-point comparisons with the baselines, and freshness limits at 7--21 days. Second, \emph{close the nutrient gap at the source}: the planned Goheung buoy is to provide nitrate and ammonium continuously, corrected for nitrite as in Section~\ref{sec:din}, and the climatology result suggests a fallback whose effect on detection must be tested before use. Third, \emph{edge deployment}: the platform plan uses the 9.2-million-parameter model as the teacher for a 3.8-million-parameter student~\citep{hinton2015distilling,gou2021knowledge}, followed by reduced numerical precision for on-site hardware; this is planned and none of it has been measured yet.

\section*{Data and service availability}
The forecasting service is operated at \url{https://oceantensor.ai.kr}. Input data come from public Korean and international providers (KOEM, KMA, KHOA, NIFS, HYCOM, Copernicus Sentinel-2 and Copernicus Marine) under their respective terms.

\section*{Author contributions}
G.K.\ developed the backbone model and the evaluation, and leads the model-compression work. C.J.\ developed the forecasting service and dashboard. Y.J.\ carried out exploratory data analysis and the IoT sensor work.

\section*{Acknowledgments}
We thank the seaweed research division of the National Institute of Fisheries Science for cooperation on bleaching case records and for advice on how recovered cases are recorded. This work was carried out within the 2026 AI Champion competition program in Korea.

\bibliographystyle{plainnat}
\bibliography{refs}

\appendix
\section{Development configurations}
\label{app:history}

Table~\ref{tab:history} lists the configurations discussed in Section~\ref{sec:history}. These runs used 2021 data from one or two tiles, small validation windows around a single event, and pixel-level confusion matrices; all but the last row predicted the day after the input window. They document the design path and are not comparable with Section~\ref{sec:results}.

\begin{table}[H]
\centering
\caption{Key development configurations (two-tile Wando+Mokpo task unless noted; F1 and precision at the recall-100\% threshold, where some source tables label the F1 as macro-F1; pixel-level counting).}
\label{tab:history}
\scriptsize
\begin{tabular}{@{}llrrrrr@{}}
\toprule
Configuration & Change & Params & ROC-AUC & PR-AUC & F1 & Precision \\
\midrule
v7 (Wando only, 30 ch) & recall-100\% threshold introduced & -- & 0.903--0.927 & -- & 0.907--0.926 & 83.3--86.2\% \\
v13 & two tiles & -- & 0.9071 & 0.8522 & 0.9193 & 87.72\% \\
v14 & + Jindo (three tiles) & -- & 0.8378 & 0.8091 & 0.7408 & 70.80\% \\
v16 & + Goheung, 2016 event (four tiles) & -- & 0.7790 & 0.7212 & 0.7580$^{\dagger}$ & -- \\
v19 & N:P ratio bug fixed; self-attention & 3,819,650 & 0.8716 & 0.8516 & 0.8953 & 84.75\% \\
xattn\_v1 & 3-group hub cross-attention & 5,922,690 & 0.9191 & 0.9110 & 0.8013 & 75.36\% \\
xattn\_v2 & 4 groups (nutrients separate) & -- & 0.9414 & -- & 0.7786 & -- \\
xattn\_v3 & residual joint + cross-attention & -- & 0.8917 & -- & 0.8953 & 84.75\% \\
xattn\_v3\_focal & + focal loss ($\gamma=2$) & -- & 0.9034 & -- & 0.9271 & 88.74\% \\
xattn\_v3\_focal\_7day & target moved to $+7$ days & -- & 0.9627 & 0.9818 & 0.8370 & 84.93\% \\
\bottomrule
\end{tabular}

\vspace{2pt}{\footnotesize $^{\dagger}$ Macro-F1 at the default threshold; the recall-100\% value was not reported for this configuration.}
\end{table}

The v13 result was later judged unreliable because it was trained before the N:P ratio bug was fixed. The first seven-day result included an external Seocheon check that was explicitly withdrawn; the next-day Seocheon check of the focal variant rests on the same seasonal pattern and should not be cited either (Section~\ref{sec:history}).

\section{Measurement provenance}
\label{app:provenance}

\begin{table}[H]
\centering
\caption{Source and date of each reported measurement.}
\label{tab:prov}
\small
\begin{tabularx}{\linewidth}{@{}Xll@{}}
\toprule
Measurement & Procedure & Date \\
\midrule
Overall metrics and confusion matrix (Table~\ref{tab:main}) & validation set, 6-seed ensemble, $\theta^\star=0.0940$ & 2026-10-04/05 \\
Same-date control (Table~\ref{tab:sdc}) & separate 5-fold CV training, within-date AUC, cluster bootstrap & 2026-10-04/05 \\
Inference port check & element-wise comparison with training path & 2026-10-04 \\
Experiments A and B (Tables~\ref{tab:expA}, \ref{tab:expB}) & ablation script with pre-registered criteria & 2026-10-06 \\
2026 patch channel audit & zero-fraction check across machines & 2026-10-07 \\
Nutrient freshness (Table~\ref{tab:gap}) & per-source latest record & 2026-10-07 \\
Nitrite share (Table~\ref{tab:no2}) & NIFS fishing-ground samples, $n=24{,}513$ & 2026-10-07 \\
Proxy search (Table~\ref{tab:nde}) & change correlations, per-tile standardized & 2026-10-07 \\
Persistence vs.\ climatology (Figure~\ref{fig:clim}) & leave-one-year-out, $n=22{,}736$ pairs & 2026-10-07 \\
\bottomrule
\end{tabularx}
\end{table}

\section{Reproduction steps}
\label{app:repro}
\begin{enumerate}
  \item Build patches: tile cube $\rightarrow$ central $64\times64$ patch per cluster and year; check the zero fraction of all 32 channels.
  \item Build labels and the split; verify the window arithmetic ($\tau-20$, length 14, horizon 7) on every sample.
  \item Train six seeds (42, 1, 2, 3, 5, 7); store normalization statistics with each checkpoint.
  \item Verify the stand-alone inference module element-wise against the training path (tolerance $10^{-4}$).
  \item Compute the overall metrics, the same-date control and the robustness experiments; compare with Tables~\ref{tab:main}--\ref{tab:expA} and \ref{tab:expB}.
  \item Run ensemble inference over all clusters and dates, build the forecast pack, and check its grade distribution against Table~\ref{tab:pack}.
  \item Deploy; the service refuses a pack whose model name does not match.
\end{enumerate}

\end{document}